\documentclass{article}

\usepackage{microtype}
\usepackage{graphicx}
\usepackage{subcaption}
\usepackage{booktabs} 

\usepackage{hyperref}

\usepackage[preprint]{icml2026}

\usepackage{amsmath}
\usepackage{amssymb}
\usepackage{mathtools}
\usepackage{amsthm}
\usepackage{multirow}

\usepackage{algorithm}
\usepackage{algorithmic}
\usepackage{bm}
\usepackage{mathtools}
\usepackage[capitalize,noabbrev]{cleveref}

\usepackage{amsmath}
\usepackage{amssymb}
\usepackage{mathtools}
\usepackage{amsthm}
\usepackage{float}
\usepackage{subcaption}
\expandafter\let\expandafter\nabla\csname nabla\endcsname

\theoremstyle{plain}

\theoremstyle{definition}

\theoremstyle{remark}

\usepackage[textsize=tiny]{todonotes}

\icmltitlerunning{PILD: Physics-Informed Learning via Diffusion}

\begin{document}

\twocolumn[
  \icmltitle{PILD: Physics-Informed Learning via Diffusion}



  \icmlsetsymbol{equal}{*}



\begin{icmlauthorlist}
\icmlauthor{Tianyi Zeng}{sjtu,equal}
\icmlauthor{Tianyi Wang}{ut,equal}
\icmlauthor{Jiaru Zhang}{sjtu}
\icmlauthor{Zimo Zeng}{zju}
\icmlauthor{Feiyang Zhang}{tju} \\
\icmlauthor{Yiming Xu}{ut}
\icmlauthor{Sikai Chen}{umn}
\icmlauthor{Junfeng Jiao}{ut}
\icmlauthor{Christian Claudel}{ut}
\icmlauthor{Xinbo Chen}{tju}
\end{icmlauthorlist}

\icmlaffiliation{sjtu}{Shanghai Jiao Tong University, Shanghai, China}
\icmlaffiliation{ut}{The University of Texas at Austin, Austin, TX, USA}
\icmlaffiliation{zju}{Zhejiang University, Hangzhou, Zhejiang, China}
\icmlaffiliation{tju}{Tongji University, Shanghai, China}
\icmlaffiliation{umn}{University of Wisconsin-Madison, Madison, WI, USA}

\icmlcorrespondingauthor{Tianyi Wang}{bonny.wang@utexas.edu}

  \icmlkeywords{Physics-Informed Neural Network, Diffusion Model}

  \vskip 0.3in
]



 \printAffiliationsAndNotice{\icmlEqualContribution}

\begin{abstract}

Diffusion models have emerged as powerful generative tools for modeling complex data distributions, yet their purely data-driven nature limits applicability in engineering and scientific problems where physical laws must be respected. 
This paper proposes Physics-Informed Learning via Diffusion (PILD), a framework that unifies diffusion modeling and physical constraints through a probabilistic residual formulation with a virtual residual observation sampled from a Laplace distribution. 
To make this formulation practical under noisy diffusion states, we introduce a Jensen-gap-aware adaptive residual scale, which reduces the bias induced by residual likelihood marginalization. 
Additionally, we develop a physics-conditional alignment mechanism for conditional tasks that encourages intermediate latent representations to remain consistent with the observation conditions during denoising. 
The proposed framework is concise, modular, and broadly applicable to problems governed by ordinary differential equations, partial differential equations, as well as algebraic equations or inequality constraints. 
Extensive experiments across engineering and scientific tasks show that PILD improves physical fidelity and predictive accuracy over representative physics-informed and diffusion-based baselines.

\end{abstract}

\section{Introduction}

Diffusion models are a rising form of probabilistic generative models that learn data distributions by adding noise to data and training neural networks to reverse this process~\cite{li2025diffusion}.
Through this iterative denoising mechanism, diffusion models have demonstrated strong performance across a wide range of domains, including computer vision~\cite{croitoru2023diffusion, liu2024residual}, temporal data modeling~\cite{ alcaraz2022diffusion}, and robust learning~\cite{blau2022threat}. 
Owing to their rigorous mathematical interpretability grounded in probability theory, diffusion models exhibit a strong capability to learn complex data distributions~\cite{yang2023diffusion}. 
As a result, diffusion models have recently achieved strong empirical performance in both scientific machine learning~\cite{ lee2023exploring} and practical engineering problems~\cite{chen2023diffusiondet}. 

On the other hand, many scientific and engineering problems impose strict physical constraints governed by underlying differential equations~\cite{lu2019deeponet}.
In such settings, training data are typically constructed either through numerical generations that enforce physical laws or through real-world measurements that inevitably contain noise~\cite{Raissi2019}.
Despite the aforementioned advances of diffusion models, samples from standard diffusion models may not always conform to physical laws, since these models primarily learn data-level distributional characteristics~\cite{huang2024diffusionpde}.
This limitation can lead to physically inconsistent predictions or reduced performance, particularly in regimes involving highly complex dynamical systems.

Some research has explored integrating physical information into generative models, for instance, by enforcing physical priors through architectural constraints~\cite{wang2025fundiff} or physics-informed loss functions \cite{huang2024diffusionpde}. 
However, many approaches introduce physical constraints as additional penalties~\cite{shan2025red, soni2025physics} or post-processing operations such as distillation \cite{sanokowski2024diffusion}, and their residual modeling often remains decoupled from the probabilistic training objective. 
Recently, \citealp{bastek2024physics} and \citealp{baldan2025flow} have taken important steps toward unifying physical models and data distributions within diffusion or flow-matching frameworks, making physics-informed generative learning a promising direction.

\begin{figure*}[htbp!]
    \centering
    \includegraphics[width=0.95\linewidth]{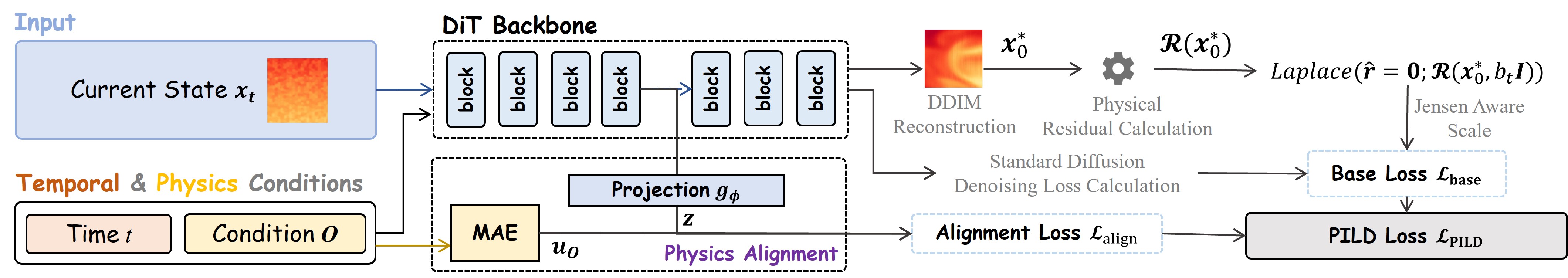}
    \caption{Overview of our PILD (Physics-Informed Learning via Diffusion) framework: The current state $\boldsymbol{x}_t$ is fed into the network, while the condition $\boldsymbol{O}$ is injected through the Physics-Conditional Alignment module. A DDIM estimate $\boldsymbol{x}_0^*$ of the clean state is then obtained for residual evaluation. By modeling the virtual residual observable $\hat{\boldsymbol{r}}$ with a Laplace distribution and using a Jensen Gap-aware effective residual scale, the framework unifies data fitting, physical regularization, and Jensen-gap-aware residual calibration.
}
    \label{fig:frame}
\end{figure*}

Based on this, we propose to embed physical laws such as partial differential equations (PDEs) directly into diffusion training through a virtual residual observation. To make this formulation practical, we further account for the bias introduced by residual likelihood marginalization under noisy diffusion states. In addition, for conditional tasks, we introduce a physics-conditional alignment mechanism that encourages intermediate latent representations of the diffusion backbone to remain semantically consistent with the given observation conditions.

Building upon these ideas, we develop a Physics-Informed Learning via Diffusion (PILD) framework (Figure~\ref{fig:frame}) that connects residual-based physics-informed learning with diffusion-based generative modeling under a unified training objective.
The main contributions are as follows:
\begin{itemize}
    \item We propose a probabilistic physics-informed diffusion framework based on Laplace residual modeling. By introducing a virtual residual observation and an adaptive Jensen-gap-aware residual scale, the method incorporates physical constraints into diffusion training while mitigating the leading-order bias caused by residual likelihood marginalization.
    \item We introduce a physics-conditional alignment strategy for conditional tasks that aligns intermediate latent representations with condition embeddings through a frozen MAE-based target representation, improving condition awareness in conditional diffusion training.
    \item We validate the proposed framework across a wide range of engineering and scientific computing benchmarks, demonstrating consistent improvements in physical fidelity and predictive performance.
\end{itemize}

\section{Related Work}
\label{related}

Diffusion models have demonstrated strong performance in generative tasks by learning to reverse a progressive noise-injection process applied to data~\cite{hang2023efficient}.
In particular, DDPM~\cite{ho2020denoising} formalized generation through a pair of Markov chains: a forward chain that perturbs data into noise, and a reverse chain that reconstructs samples from noise.
Since then, diffusion models have achieved significant progress across various fields, especially in processing simple time series signals~\cite{ yi2024time}.
For instance, AEDM~\cite{fu2025diffusion} applied diffusion models for denoising acoustic emission signals in concrete damage detection.
Models like DDIM~\cite{song2020denoising} and EDM~\cite{karras2022elucidating} offered expressive generative capabilities but remained limited when sampling with specific physical constraints. 

Motivated by these limitations, recent work has attempted to incorporate physical priors into diffusion processes~\cite{zhang2025physics, yang2025pis}.
Notably, PIDM~\cite{bastek2024physics} unified denoising diffusion models with PINNs by incorporating PDE constraints into the training objective, reducing the residual errors of generated samples compared to the model proposed by~\citealp{shu2023physics} and CoCoGen~\cite{jacobsen2025cocogen}.
Building upon~\citealp{shu2023physics}'s model, Pi-Fusion~\cite{qiu2024pi} was designed for learning fluid dynamics based on diffusion models.
DPS~\cite{shi2024diffusion} introduced a novel computational technique for solving PDEs via PINN in the context of diffusion-based sampling.
Inspired by Neural Network diffusion~\cite{wang2024neural}, \citealp{cheng2025diffpinn} presented a latent diffusion-based approach to efficiently initialize PINNs for seismic wavefield modeling.
Other studies like RED-DiffEq~\cite{shan2025red} integrated diffusion models directly into PDE-governed inverse problems as a regularization mechanism.
Despite this progress, existing physics-informed diffusion models can still produce physically inconsistent solutions because of limited residual modeling or insufficient physics-conditioned guidance. 
Therefore, the full potential of approaches integrating PINNs with diffusion models remains largely unexplored.

\section{Methodology}
\label{Proposed}

The objective of this work is to improve the physical fidelity of diffusion-based generative models while preserving their ability to learn complex data distributions. To this end, we incorporate physical constraints into diffusion training through a probabilistic residual likelihood. Different from existing physics-informed diffusion frameworks with fixed or timestep-coupled penalties, our method uses a Laplace residual model with a Jensen-gap-aware adaptive residual scale to account for residual marginalization bias under noisy latent states. For conditional tasks, we further introduce a physics-conditional alignment mechanism to enhance consistency with the observation conditions. The preliminaries of diffusion models and PINNs are provided in Appendix \ref{sup:pre}.

\subsection{Physics-Informed Diffusion Training Framework}

Our method assumes access to training data sampled either from simulators or real-world measurements. 
This assumption is fundamental: if training data contain substantial errors or fundamentally violate the objective physical laws, the resulting data distribution will fail to reflect the desired physics, thereby misleading the learning process. 
In practice, although real-world measurements may contain noise or outliers, they generally conform to the correct physical principles in a probabilistic sense, making them suitable for physics-informed learning. 
Consequently, system identification lies outside the scope of this work.

Our approach relies on estimating the clean sample $\boldsymbol{x}_0 \sim q(\boldsymbol{x}_0,\boldsymbol{O})$ from a noisy state $\boldsymbol{x}_t$ using the network prediction $\boldsymbol{\epsilon}_\theta$ under condition $\boldsymbol{O}$. 
According to \cite{ho2020denoising}, diffusion training maximizes a variational lower bound. The data-driven component is therefore
\begin{equation}
    \arg \max_{\boldsymbol{\theta}} \mathbb{E}_{\boldsymbol{x}_0 \sim q(\boldsymbol{x}_0,\boldsymbol{O})} [\log p_{\boldsymbol{\theta}}(\boldsymbol{x}_0 | \boldsymbol{O})].
\end{equation}

For the physics-informed component, we introduce a virtual residual observation $\hat{\boldsymbol{r}}=\mathbf{0}$ and model its likelihood with a Laplace distribution:
\begin{equation}
q_{\boldsymbol{\mathcal{R}}}(\hat{\boldsymbol{r}}|\boldsymbol{x}_0)
=
{Laplace}(\hat{\boldsymbol{r}}; \boldsymbol{\mathcal{R}}(\boldsymbol{x}_0), \sigma \boldsymbol{I}),
\end{equation}
where $\boldsymbol{\mathcal{R}}(\boldsymbol{x}_0)$ denotes the physical residual operator. 
This provides a probabilistic interpretation of residual minimization while remaining more robust to outliers than Gaussian modeling \cite{kotz2012laplace}. 
The associated virtual likelihood is
\begin{equation}
\begin{aligned}
p_\theta(\hat{\boldsymbol{r}}) 
&= \int p_\theta(\hat{\boldsymbol{r}}, \boldsymbol{x}_0) \, \mathrm{d}\boldsymbol{x}_0 = \int q_{\mathcal{R}}(\hat{\boldsymbol{r}}| \boldsymbol{x}_0) p_\theta(\boldsymbol{x}_0,\boldsymbol{O}) \, \mathrm{d}\boldsymbol{x}_0 \\
&= \mathbb{E}_{\boldsymbol{x}_0 \sim p_\theta(\boldsymbol{x}_0,\boldsymbol{O})} \left[ q_{\mathcal{R}}(\hat{\boldsymbol{r}}| \boldsymbol{x}_0) \right].
\end{aligned}
\end{equation}
Accordingly, the exact physics-informed contribution is
\begin{equation}
\label{eq:exact_phys_obj_main}
\mathbb{E}_{\hat{\boldsymbol{r}}}\!\left[
\log \mathbb{E}_{\boldsymbol{x}_0 \sim p_\theta(\boldsymbol{x}_0,\boldsymbol{O})}
\left[q_{\mathcal{R}}(\hat{\boldsymbol{r}}|\boldsymbol{x}_0)\right]
\right].
\end{equation}

Combining this term with the data likelihood yields a joint objective over observed data and virtual residual observations. 
As discussed in Appendix \ref{A1}, this formulation remains compatible with the score-based interpretation of diffusion models in the idealized case where the target data distribution satisfies the governing physical constraints.

However, the exact objective in Equation \eqref{eq:exact_phys_obj_main} is intractable in practical diffusion training, since it requires marginalizing the residual likelihood over latent clean states drawn from the model posterior. 
To obtain a tractable surrogate, we condition on $(\boldsymbol{x}_t,\boldsymbol{O})$ and approximate the clean state by a denoised representative $\boldsymbol{x}_0^*$. 
This gives rise to two distinct approximation gaps \cite{gao2017bounds}.

The first gap is the same type discussed in \citealp{bastek2024physics, baldan2025flow, zhang2025physics}. 
The residual is evaluated on a single representative of the clean state rather than marginalized over the full posterior clean-state distribution. 
If one directly uses the conditional mean, then for a nonlinear residual operator generally has
\begin{equation}
\boldsymbol{\mathcal{R}}\!\left(\mathbb{E}[\boldsymbol{x}_0| \boldsymbol{x}_t,\boldsymbol{O}]\right)
\neq
\mathbb{E}\!\left[\boldsymbol{\mathcal{R}}(\boldsymbol{x}_0)| \boldsymbol{x}_t,\boldsymbol{O}\right].
\end{equation}
Following \citealp{bastek2024physics}, we use DDIM to estimate $\boldsymbol{x}_0^*$ from $\boldsymbol{x}_t$ and evaluate the residual on $\boldsymbol{x}_0^*$, which is closer than directly using the conditional mean to evaluating the residual on actual samples.

The second gap arises from the probabilistic structure of our physics-informed objective itself. 
Even if the posterior over clean states were available, Equation \eqref{eq:exact_phys_obj_main} still contains a logarithm of an expectation over latent clean states, which in general cannot be replaced exactly by an expectation of a logarithm or by a residual penalty evaluated at a single representative. 
This is the more direct Jensen-type bias introduced by residual likelihood marginalization.

We first define a timestep-dependent base scale
\begin{equation}
\label{eq:bt_main}
b_t = B_t / c,
\end{equation}
where $B_t$ is the fixed variance of the denoising process and $c>0$ controls the overall strength of the physical penalty.

Under the Laplace model, the exact negative physical log-likelihood conditioned on $(\boldsymbol{x}_t,\boldsymbol{O})$ becomes
\begin{equation}
\label{eq:true_free_energy_main}
\mathcal{L}^{\mathrm{true}}_{\mathrm{phys}}(t)
=
-\log
\mathbb{E}
\left[
\exp\!\left(-\frac{\|\boldsymbol{\mathcal{R}}(\boldsymbol{x}_0)\|_1}{b_t}\right)
\,\middle|\,
\boldsymbol{x}_t,\boldsymbol{O}
\right].
\end{equation}
This term is the exact conditional free energy associated with the virtual residual likelihood. 
To derive a practical loss of the same positive residual-penalty form used in diffusion training, we next approximate this quantity while keeping track of the bias induced by the log-marginalization.
Let
\begin{equation}
\begin{aligned}
&S_t := \|\boldsymbol{\mathcal{R}}(\boldsymbol{x}_0)\|_1,
\\
&\mu_t := \mathbb{E}[S_t \mid \boldsymbol{x}_t,\boldsymbol{O}],
\qquad
\sigma_t^2 := \mathrm{Var}(S_t \mid \boldsymbol{x}_t,\boldsymbol{O}).
\end{aligned}
\end{equation}
As shown in Appendix \ref{A1}, a second-order expansion gives
\begin{equation}
\label{eq:second_order_phys_main}
\mathcal{L}^{\mathrm{true}}_{\mathrm{phys}}(t)
\approx
\frac{\mu_t}{b_t}
-
\frac{\sigma_t^2}{2b_t^2},
\end{equation}
so that, relative to the naive first-order surrogate $\mu_t / b_t$, the leading-order log-marginalization bias is
\begin{equation}
\label{eq:jensen_gap_main}
\Delta_t
\approx
\frac{\sigma_t^2}{2b_t^2}.
\end{equation}
This suggests that the effective residual weighting should depend not only on the diffusion-induced base scale $b_t$, but also on the conditional residual variance.

To account for this bias while retaining the same positive residual-penalty form, we introduce an effective residual scale by matching the first-order surrogate to the second-order expansion as
\begin{equation}
\label{eq:effective_bt_main}
\tilde b_t
=
b_t
+
\frac{\sigma_t^2}{2\mu_t+\varepsilon},
\end{equation}
where $\varepsilon>0$ is a small constant for numerical stability, the detailed derivation is given in Appendix \ref{A1}.
In practice, the conditional moments $\mu_t$ and $\sigma_t^2$ are not directly available. 
We therefore approximate them by timestep-wise minibatch plug-in statistics computed from the DDIM-based residual magnitudes, and stabilize these estimates using exponential moving averages (EMA) over training iterations. 
For a minibatch $\mathcal{B}_t$ at timestep $t$, let
\begin{equation}
s_i = \left\|\boldsymbol{\mathcal{R}}(\boldsymbol{x}_{0,i}^*)\right\|_1, \qquad i \in \mathcal{B}_t,
\end{equation}
and define
\begin{equation}
\label{eq:mu_sigma_est}
\hat{\mu}_t = \frac{1}{|\mathcal{B}_t|}\sum_{i\in\mathcal{B}_t} s_i,
\qquad
\hat{\sigma}_t^2 = \frac{1}{|\mathcal{B}_t|-1}\sum_{i\in\mathcal{B}_t}(s_i-\hat{\mu}_t)^2.
\end{equation}
Using the corresponding EMA-smoothed statistics $\bar{\mu}_t$ and $\bar{\sigma}_t^2$, we set
\begin{equation}
\label{eq:effective_bt_hat_main}
\hat{\tilde b}_t
=
b_t
+
\frac{\bar{\sigma}_t^2}{2(\bar{\mu}_t+\varepsilon)},
\end{equation}
which yields the practical physics loss
\begin{equation}
\label{eq:phys_loss_main}
\mathcal{L}_{\mathrm{phys}}(t)
=
\frac{1}{\hat{\tilde b}_t}
\left\|
\boldsymbol{\mathcal{R}}(\boldsymbol{x}_0^*)
\right\|_1.
\end{equation}
This introduces little overhead beyond residual evaluation itself. Here, DDIM provides a practical clean-state representative rather than an exact posterior sample, while $\hat{\tilde b}_t$ supplies a timestep-adaptive correction for the leading-order bias induced by residual likelihood log-marginalization. In addition, the diffusion-induced base scale $b_t$ acts as an intrinsic stabilizer by lower-bounding the effective residual scale and reducing the relative influence of noisy minibatch variance estimates. The detailed EMA update rule is provided in Appendix \ref{A2}.

During optimization, $p_{\theta}$ gradually approaches $q$. We therefore formulate the physics-informed base loss under the same expectation over data and noisy states (see Appendix \ref{A4}) as
\begin{equation}
\begin{aligned}
\label{eq:total_loss_unified_robust}
&\mathcal{L}_{\text{base}}(\boldsymbol{\theta})
=
\mathbb{E}_{t \sim [1,T], (\boldsymbol{x}_0, \boldsymbol{O}) \sim q(\boldsymbol{x}_0, \boldsymbol{O}), \boldsymbol{\epsilon} \sim \mathcal{N}(\mathbf{0}, \mathbf{I})}
\\
&\Big[
\lambda_t \| \boldsymbol{\epsilon} - \boldsymbol{\epsilon}_\theta \|^2 
+
\frac{1}{\hat{\tilde b}_t}
\left\|
\boldsymbol{\mathcal{R}}({\boldsymbol{x}}_0^*)
\right\|_1
\Big],
\end{aligned}
\end{equation}
where $\lambda_t$ is a time-dependent Min-SNR weighting \cite{hang2023efficient}. Compared with existing physics-informed diffusion formulations that use fixed or purely timestep-coupled scales, our formulation addresses two approximation sources in a unified manner: DDIM-based clean-state estimation mitigates the posterior-representative gap discussed in PIDM, while the adaptive residual scale accounts for the leading-order bias induced by residual likelihood log-marginalization.

This framework can also be extended to other types of constraints; the corresponding formulations are provided in Appendix \ref{app:otherform}.

\subsection{Physics-Conditional Alignment}

For each benchmark, the noisy input $\boldsymbol{x}_t$ is tokenized according to its native modality, using a 1D patch embedding for sequence data and a 2D patch embedding for field data. The resulting tokens are processed by Transformer blocks, while the diffusion timestep and observation condition $\boldsymbol{O}$ are injected through adaptive layer normalization, following the standard conditioning mechanism of DiT.

For conditional tasks, the observation $\boldsymbol{O}$ is first encoded into a compact embedding and used to modulate all DiT blocks. To further encourage the latent dynamics to remain consistent with the conditioning observation, we introduce a physics-conditional alignment regularizer. Let $\boldsymbol{H}^{(l)} \in \mathbb{R}^{N \times d}$ denote the token features at the output of the $l$-th DiT block. We pool them into a global representation
\begin{equation}
\bar{\boldsymbol{h}}^{(l)} = \mathrm{Pool}(\boldsymbol{H}^{(l)}),
\end{equation}
and map it through a lightweight projection head $g_{\phi}$:
\begin{equation}
\boldsymbol{z}^{(l)} = g_{\phi}(\bar{\boldsymbol{h}}^{(l)}).
\end{equation}

We then introduce a frozen task-specific MAE encoder $f_{\mathrm{MAE}}$ trained on the corresponding observations, and use its output
\begin{equation}
\boldsymbol{u}_{\boldsymbol{O}} = f_{\mathrm{MAE}}(\boldsymbol{O})
\end{equation}
as the alignment target. The alignment loss is defined by cosine similarity:
\begin{equation}
\mathcal{L}_{\mathrm{align}}^{(l)}
=
1
-
\frac{
\left\langle
\boldsymbol{z},
\mathrm{sg}\!\left[\boldsymbol{u}_{\boldsymbol{O}}\right]
\right\rangle
}{
\|\boldsymbol{z}\|_2
\,
\|\mathrm{sg}\!\left[\boldsymbol{u}_{\boldsymbol{O}}\right]\|_2
},
\end{equation}
where $\mathrm{sg}[\cdot]$ denotes stop-gradient. 
The final training objective combines the physics-informed base loss and the alignment regularizer:
\begin{equation}
\label{eq:pild_final_loss}
\mathcal{L}_{\mathrm{PILD}}
=
\mathcal{L}_{\mathrm{base}}
+
\lambda_{\mathrm{align}} \mathcal{L}_{\mathrm{align}}.
\end{equation}
For unconditional tasks, the alignment term is omitted. In practice, we additionally employ early stopping based on the validation loss to avoid late-stage training instability \cite{yu2024representation, wang2026repa}.

\subsection{Diffusion Inference and Sampling}

Once the model $\boldsymbol{\epsilon}_\theta$ has been trained by minimizing $\mathcal{L}_{\text{PILD}}$, it can be used as a conditional generator to produce new solutions. Following \cite{bastek2024physics}, we adopt a two-step DDIM sampler at inference for computational efficiency. Starting from $\boldsymbol{x}_T \sim \mathcal{N}(\mathbf{0}, \mathbf{I})$, the sampler first maps $\boldsymbol{x}_T$ to an intermediate state near $t=1$ and then reconstructs the final clean sample $\boldsymbol{x}_0$. The detailed procedure is provided in Appendix \ref{A5_sampling}.

By manipulating the input conditions $\boldsymbol{O}$, we can perform ``what-if'' analyses and generate corresponding solution profiles. Because physical constraints are incorporated during training, the sampled outputs tend to exhibit improved physical plausibility compared with purely data-driven diffusion models, while still reflecting the diversity of the learned conditional data distribution. This generative capability enables not only point prediction but also uncertainty characterization via sample ensembles under fixed conditions.

\section{Experiments}
\label{experiments}

To evaluate whether PILD provides stable physical guidance and improves physical consistency during training, we conduct experiments on four datasets covering engineering systems and scientific machine learning.
Due to space limitations, details for each experiment, baseline methods, and ablations are provided in Appendix \ref{A5} and \ref{A6}; only key settings and results are presented here.
For fairness, all compared methods use the same train/test splits, input variables, preprocessing pipeline, normalization factors, and evaluation metrics within each benchmark. Baselines are selected to cover classical engineering estimators, deterministic neural predictors, PINN-style models, and diffusion or flow-based generative models, so the comparisons are intended to evaluate both task-specific accuracy and the effect of physics-informed generative modeling under matched data conditions.
When $\pm$ values are reported in tables, they denote the standard deviation across different generated samples or evaluated trajectories within a single trained-model evaluations.
In the tables, bold values indicate the best performance and underlined values indicate the second-best performance.

\subsection{Vehicle Tracking}

\begin{figure}[ht!]
    \centering
    \includegraphics[width=0.95\linewidth]{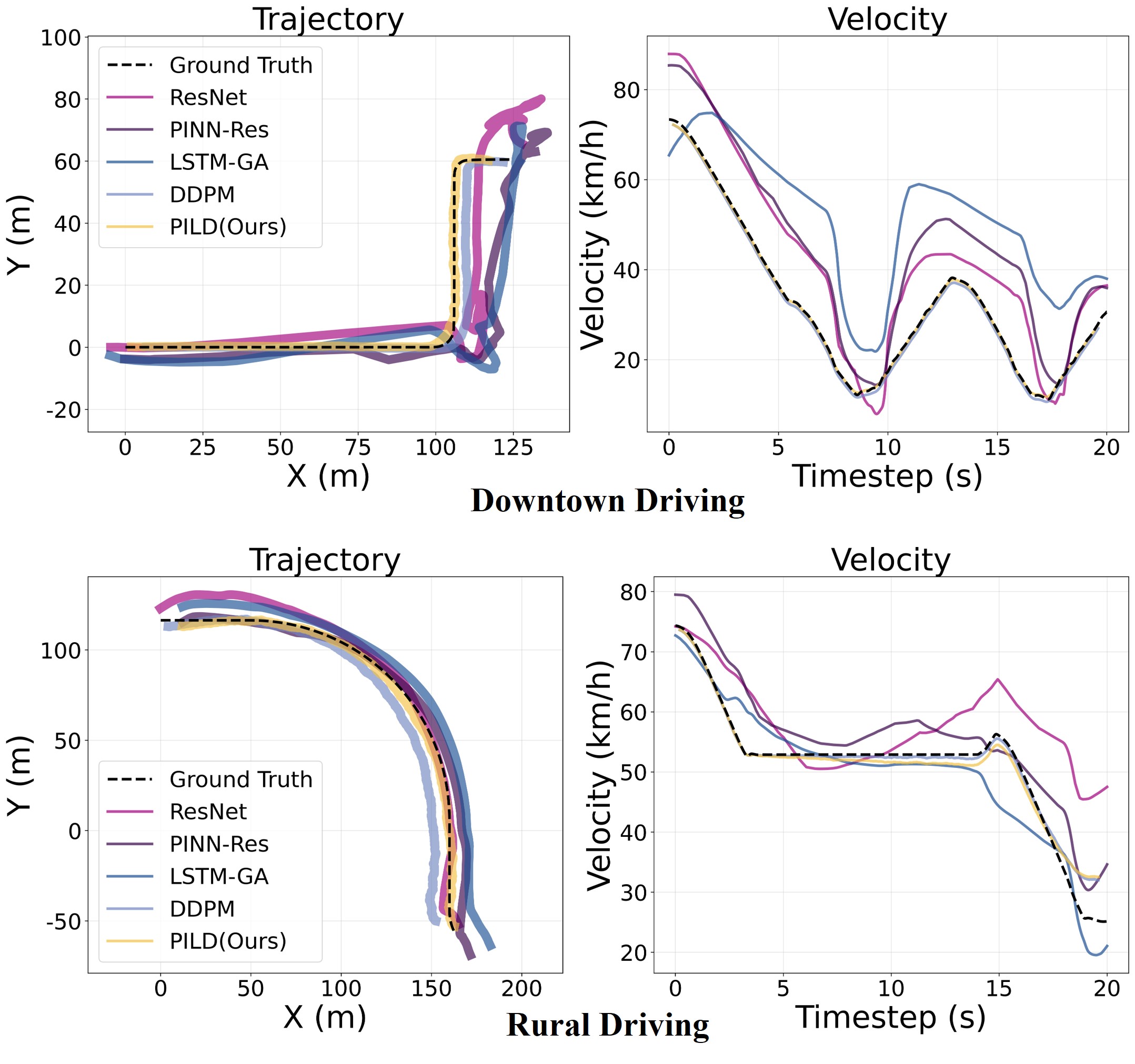}
    \caption{Experiment results on tracking tasks.}
    \label{fig:result_traj}
\end{figure}

\begin{table}[htbp!]

\begin{center}
\setlength{\tabcolsep}{3pt} 

\footnotesize
\begin{tabular}{l|ccc|ccc}
\toprule
\multirow{2}*{\textbf{Method}} & \multicolumn{3}{|c|}{\textbf{Downtown Driving}} & \multicolumn{3}{|c}{\textbf{Rural Driving}} \\ 
~ & \textbf{$e_{x\&y}$$\downarrow$}          & \textbf{$e_\psi$$\downarrow$}          & \textbf{$e_{v_x}$$\downarrow$}    & \textbf{$e_{x\&y}$$\downarrow$}          & \textbf{$e_\psi$$\downarrow$}          & \textbf{$e_{v_x}$$\downarrow$}\\

\midrule

EKF &  6.602 &  6.930 & 5.476 &  6.179 & 6.241& 5.512 \\
FCN &  9.567 &  7.977 & 14.284 &  9.623 & 6.606& 10.972 \\

ResNet & 9.236 & 6.899 & 9.094 & 5.517 &10.361& 9.315 \\

PINN-FCN & 8.824 & 14.488& 17.287 &8.279 &10.349&15.251  \\

PINN-ResNet  &5.923 &9.753 & 11.286 & \underline{3.729} & 9.134 & 8.248\\

B-PINN  & \underline{5.867} &6.597 & 10.563 & 6.034 & 6.511 & 9.656 \\

LSTM-RNN & 8.953 & 10.199 & 11.298 & 9.643 &8.831& 8.168 \\
LSTM-GA & 9.019 & 6.579 & 8.442 & 9.005 & \underline{5.994} & 10.837 \\

DDPM & 9.653 & \underline{6.109} & \underline{4.317} & 4.697 &7.138& \underline{4.936}\\

\textbf{PILD} (Ours) & \textbf{4.796} &  \textbf{5.391} &  \textbf{3.048} & \textbf{3.562} &  \textbf{5.837} &  \textbf{3.033}  \\ \bottomrule

\end{tabular}
\caption{Quantitative results on tracking tasks.}
\label{table:traj}
\end{center}
\end{table}

Vehicle tracking requires a vehicle dynamic model to predict the future state. The Ackerman steering model \cite{mitchell2006analysis} includes longitudinal, lateral, and yaw motions, which can be expressed as:
\begin{equation}
[x_{t+1},y_{t+1},\psi_{t+1},v_{x_{t+1}}]^\top
    =
    \boldsymbol{\mathcal{F}}([x_{t},y_{t},\psi_{t},v_{x_{t}}]^\top,[\theta,\dot{u}]^\top),
\end{equation}
where $[x,y,\psi,v]^\top$ is the vehicle state vector ($x$ coordinate, $y$ coordinate, yaw angle, forward velocity), $[\theta,\dot{u}]^\top$ is the control vector ($\theta$ steering angle, $\dot{u}$ acceleration), and the system transfer equation is denoted as $\boldsymbol{\mathcal{F}}$.

This task aims to predict the vehicle state at the next time step based on its state and control input at the current time step.
The methods used for comparison include \textbf{EKF} \cite{ribeiro2004kalman}, \textbf{FCN} \cite{long2015fully},  \textbf{ResNet} \cite{he2016deep}, \textbf{PINN-FCN} \cite{Long2022}, \textbf{PINN-ResNet} \cite{Long2022}, \textbf{B-PINN} \cite{EKI-B-PINNs}, \textbf{LSTM-RNN} \cite{ip2021vehicle}, \textbf{LSTM-GA} \cite{zeng2024wheel} and \textbf{DDPM} \cite{ho2020denoising}.
The experiments are performed on a vehicle driving dataset \cite{wang2025rad} with two scenarios: downtown and rural driving. 
Trajectory prediction is performed incrementally, where each predicted point is fed into the next step, to evaluate error accumulation over time under the same protocol for all methods.

The quantitative results are presented in Table \ref{table:traj}.
Due to space constraints, the results with sample-level standard deviations are presented in Table \ref{table:app:tracking} of Appendix \ref{app:tracking}.
We select one trajectory for each scenario for demonstration in Figure \ref{fig:result_traj}. 
Basic methods and LSTM-based methods perform less effectively on the mixed-vehicle dataset, as they must handle variations in data characteristics caused by different physical model parameters. PINNs are also constrained by fixed physical models, which can limit their adaptability across vehicle types. Consequently, both categories exhibit error accumulation over time, leading to trajectory deviations, as shown in Figure \ref{fig:result_traj}. In contrast, diffusion-based methods better capture the underlying data distribution under the same inputs, resulting in lower prediction errors in this setting.

\subsection{Tire Forces}

The measurement of vehicle dynamic tire forces has long been a focal topic within the automotive community, which directly govern both safety and dynamic stability of vehicles \cite{zeng2024analysis}. 
The vertical tire force serves as the foundation for tire forces in other directions. 
Therefore, it is typically of primary interest and can be expressed as:
\begin{equation}
\frac{\partial F_z(t,T)}{\partial t} = \boldsymbol{\mathcal{{F}}}(\boldsymbol{O}),
\end{equation}
where $\boldsymbol{O}$ is the set of sensors' data, $t$ is time, $T$ is tire temperature, and $\boldsymbol{\mathcal{F}}$ denotes the tire force calculation term.

We evaluate our approach on a racing car chassis dynamic dataset \cite{zeng2024wheel} with three sets of working conditions, including aggressive, sporty, and smooth driving.
The methods used for comparison include \textbf{EKF} \cite{ribeiro2004kalman},
\textbf{PINN-FCN} \cite{Long2022}, \textbf{PINN-ResNet} \cite{Long2022}, \textbf{B-PINN} \cite{EKI-B-PINNs}, \textbf{LSTM-RNN} \cite{ip2021vehicle}, \textbf{LSTM-GA} \cite{zeng2024wheel}, \textbf{DDPM} \cite{ho2020denoising}, and \textbf{PIDM} \cite{bastek2024physics}.

The quantitative results are presented in Table \ref{table:wheelload}.
The results with sample-level standard deviations are also provided in Table \ref{table:app:tire} of Appendix \ref{tireforce}.
We select three methods for demonstration in Figure \ref{fig:result_wheelload0}; more samples are provided in Figure \ref{fig:result_wheelload} in Appendix \ref{tireforce}.
Since all required sensors are installed in the chassis, the collected data contain significant noise.
Therefore, we conduct ablation studies on whether to apply denoising to the input. As presented in Appendix \ref{ablation}, the experimental results highlight the importance of data preprocessing and evaluate the robustness of our method under noisy inputs.
Different vehicle types have different chassis dynamics, posing challenges to physical consistency.
While the EKF is a classical method for addressing such problems, it underperforms learning-based approaches when the dataset includes multiple working conditions and inherent noise. Diffusion-based methods achieve competitive performance, with physics-informed diffusion models further reducing the error. These results support the usefulness of Laplace-distributed residual modeling for practical engineering data.

\begin{figure}[t!]
    \centering
    \includegraphics[width=1\linewidth]{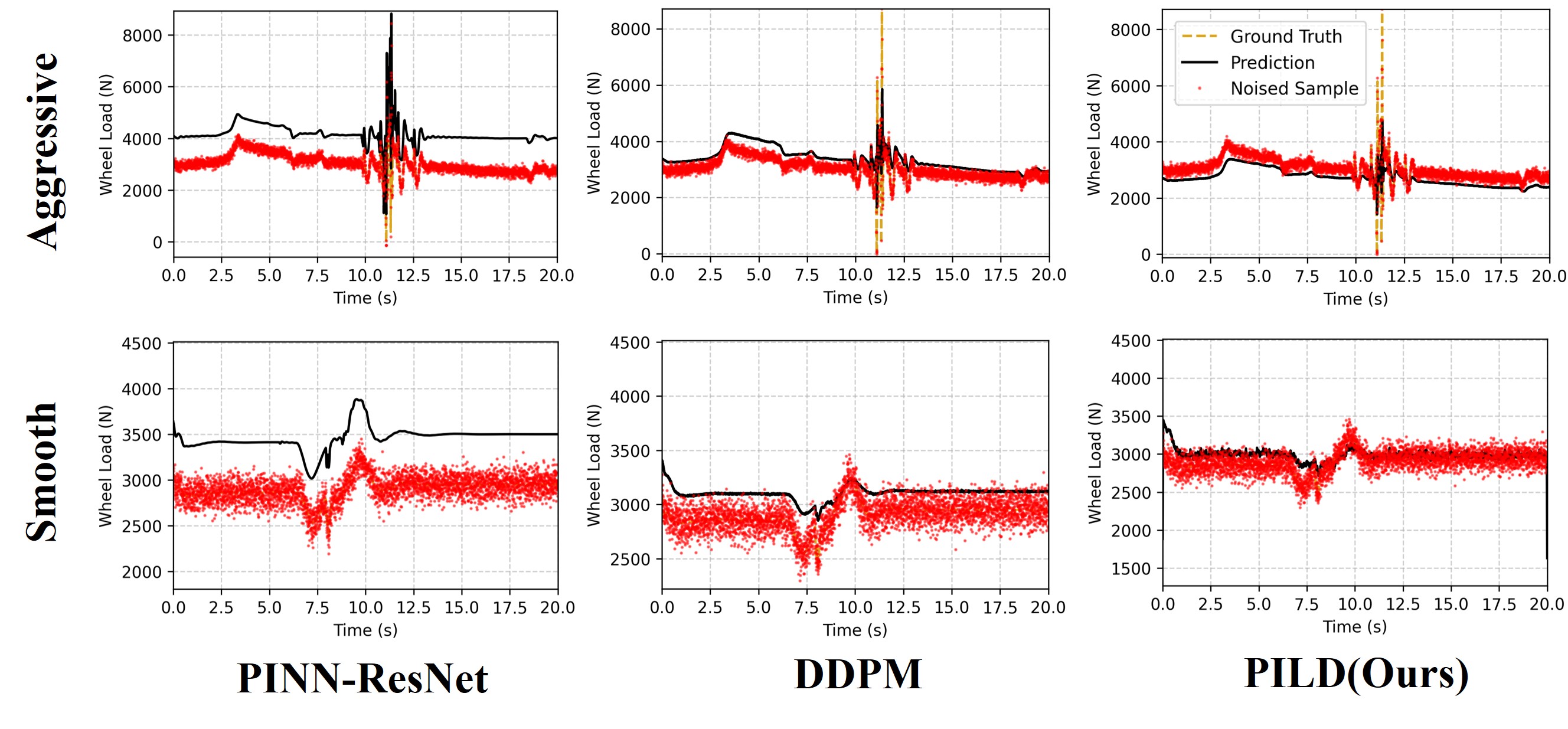}
    \caption{Samples of tire force estimation experiments on PINN, DDPM and PILD (Ours).}
    \label{fig:result_wheelload0}
\end{figure}

\begin{table}[t!]
\begin{center}
\setlength{\tabcolsep}{10pt} 
\footnotesize
{
\begin{tabular}{l|c|c|c}
\toprule
\multirow{2}*{\textbf{Method}} & \multicolumn{1}{|c|}{\textbf{Aggressive}} & \multicolumn{1}{|c}{\textbf{Smooth}} & \multicolumn{1}{|c}{\textbf{Sporty}} \\ 
~ & \textbf{$e_{F}$$\downarrow$}          & \textbf{$e_F$$\downarrow$}          & \textbf{$e_F$$\downarrow$}   \\

\midrule

EKF &  1008.127&  597.363 & 716.543 \\

PINN-FCN & 1012.849 & 551.508 & 654.644 \\

PINN-ResNet  &994.783 &642.110 & 653.944 \\

B-PINN  &1002.531 &598.617& 661.279 \\

LSTM-RNN & 1302.589 & 720.567 & 788.163 \\
LSTM-GA & 1150.434 & 582.840 & 642.638\\

DDPM & \underline{980.328} & \underline{560.796} & 660.897\\

PIDM & 988.642 & 579.546 & \underline{630.543} \\

\textbf{PILD} (Ours) &  \textbf{958.578} &  \textbf{520.631} & \textbf{607.274} \\ \bottomrule

\end{tabular}
\caption{Quantitative results on tire force estimation tasks.}
\label{table:wheelload}
}
\end{center}
\end{table}

\subsection{Darcy Flow}
Modeling subsurface flow requires solving a parameterized partial differential equation governed by Darcy's law \cite{hubbert1956darcy}. The relationship between the permeability field $k(\boldsymbol{x})$ and the resulting pressure field $p(\boldsymbol{x})$ is expressed as:
\begin{equation}
    -\nabla \cdot (k(\boldsymbol{x}) \nabla p(\boldsymbol{x})) = f(\boldsymbol{x}), \quad \boldsymbol{x} \in \Omega,
\end{equation}
where $\Omega$ denotes the spatial domain, $f(\boldsymbol{x})$ is a known source term, and appropriate boundary conditions are imposed on $\partial\Omega$. The model's output consists of two channels, corresponding to the permeability field $k$ and the pressure field $p$, and the PDE residual $\mathcal{R}(k, p) = -\nabla \cdot (k \nabla p) - f$ is incorporated directly into the training objective to enforce physical consistency.

We follow the common generative benchmark protocol used in prior work and model the joint field distribution of permeability and pressure. The model generates paired fields without additional sparse observation conditioning at inference.
The Darcy flow dataset used in our experiments is from \citealp{jacobsen2025cocogen}.
The methods used for comparison include \textbf{DDPM} \cite{ho2020denoising}, \textbf{FNO} \cite{li2020fourier}, \textbf{DiffusionPDE} \cite{huang2024diffusionpde}, \textbf{CoCoGen} \cite{jacobsen2025cocogen}, \textbf{PBFM} \cite{baldan2025flow}, and \textbf{PIDM} \cite{bastek2024physics}.

\begin{figure}[t!]
    \centering
    \includegraphics[width=1.01\linewidth]{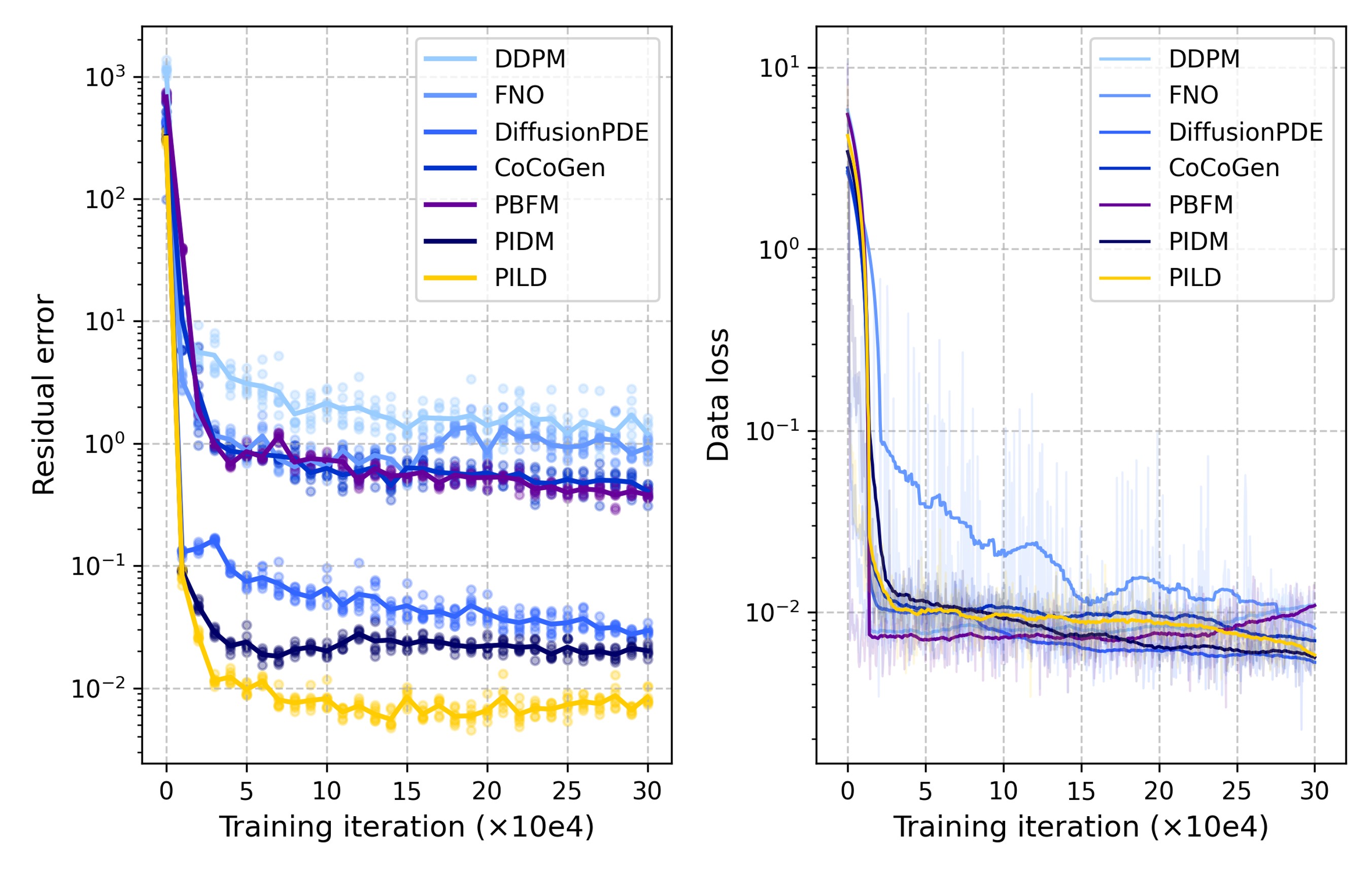}
    \caption{Evaluation of residual error and data loss of Darcy flow.}
    \label{fig:darcy1}
\end{figure}

\begin{figure}[t!]
    \centering
    \includegraphics[width=1\linewidth]{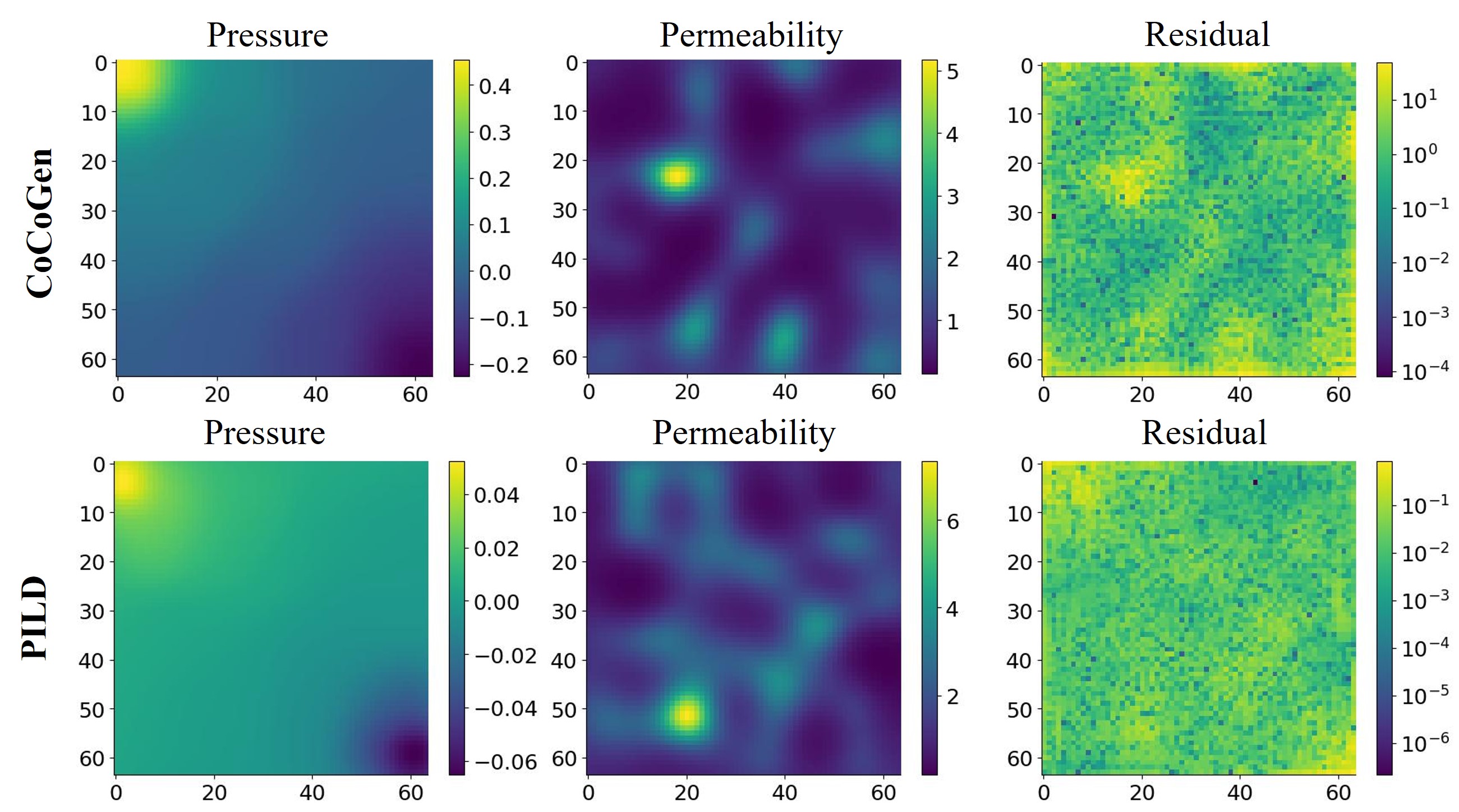}
    \caption{Samples of Darcy flow generation experiments on CoCoGen and PILD (Ours).}
    \label{fig:darcy3}
\end{figure}

The testing data loss and residual error are shown in Figure \ref{fig:darcy1}. 
In terms of physical residual error, PILD obtains a lower residual than the compared baselines, supporting the effectiveness of the Laplace-distributed residual modeling. 
Regarding data loss, as the Darcy flow task is unconditional, generation is constrained only by the unified loss. PILD reduces physical residual error while maintaining comparable data loss, suggesting that the physical improvement does not come from simply sacrificing data consistency.
We select several samples for demonstration in Figure \ref{fig:darcy3}.
More samples are provided in Figure \ref{fig:darcy2} in Appendix \ref{darcy}.

\subsection{Plasma Dynamics}

Modeling edge plasma turbulence requires solving the drift-reduced Braginskii equations \cite{mathews2021uncovering}, a system of nonlinear PDEs governing the evolution of key plasma fields. Of particular interest are the electron density $n(\boldsymbol{x}, t)$ and the electron temperatures $T_e(\boldsymbol{x}, t)$, whose normalized logarithmic forms evolve according to:
\begin{equation}
\begin{aligned}
    \frac{d \ln n}{dt} = \mathcal{S}_n(n, \phi, T_e, \ldots), 
    \frac{d \ln T_e}{dt} = \mathcal{S}_{T_e}(n, T_e, \phi, \ldots), 
\end{aligned}
\end{equation}

\begin{figure}[t!]
    \centering
    \includegraphics[width=1\linewidth]{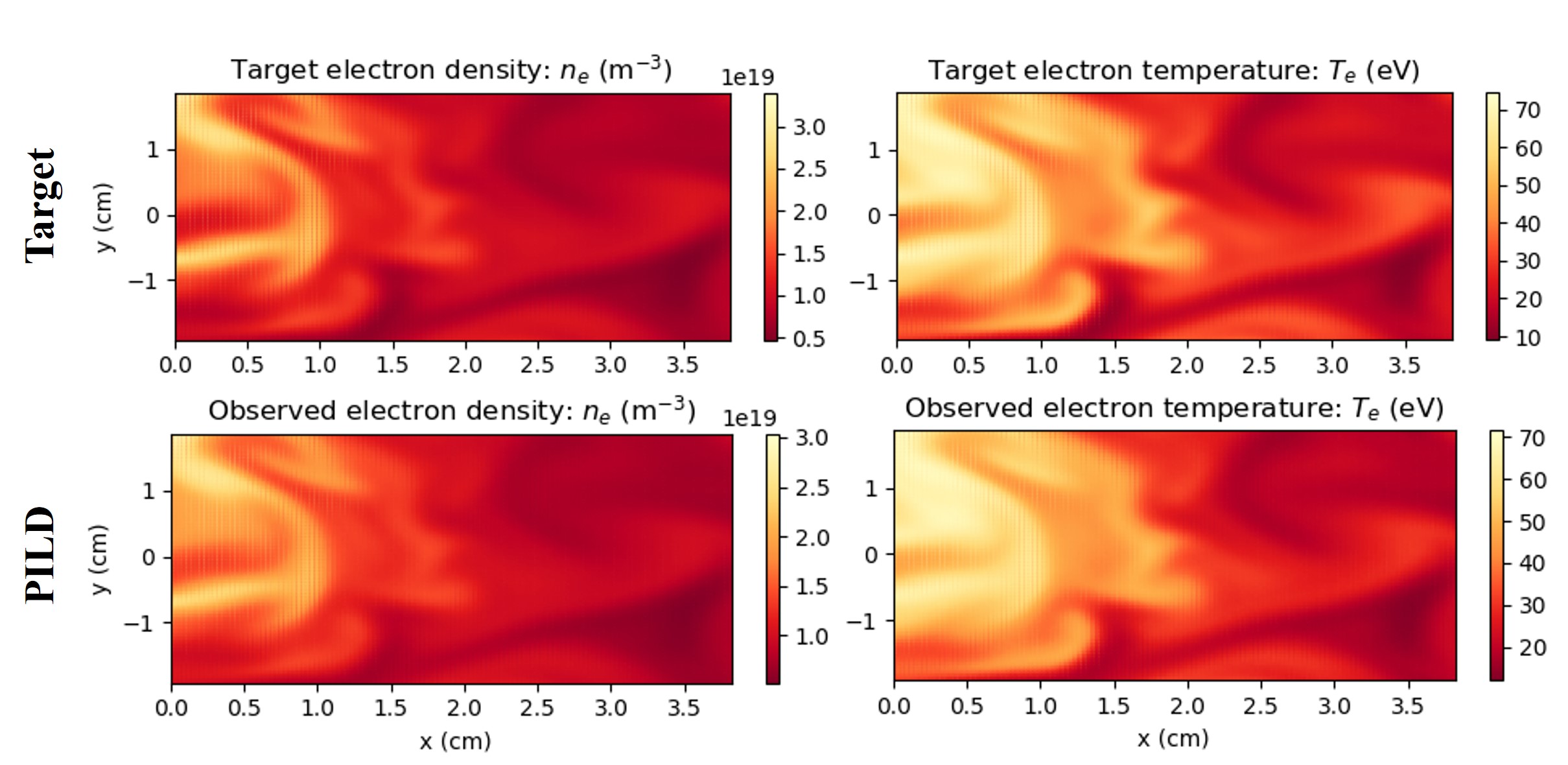}
    \caption{Samples of plasma dynamics prediction experiments on PILD (Ours).}
    \label{fig:plasma0}
\end{figure}

\begin{table}[t!]
\begin{center}
\setlength{\tabcolsep}{5pt} 
\footnotesize
{
\begin{tabular}{l|c|c}
\toprule
\multirow{1}*{\textbf{Method}} & Density error ($10^{19}$) & Temperature error \\  
\midrule

PINN & 0.531 $\pm$ 0.095 & 5.292 $\pm$ 0.538\\
DDPM  & 0.122 $\pm$ 0.020 & 1.766 $\pm$ 0.284\\
FNO & 0.153 $\pm$ 0.046 & 1.609 $\pm$ 0.258\\
DiffusionPDE & 0.112 $\pm$ 0.014 & \underline{0.728 $\pm$ 0.067}\\

PBFM  & {0.127 $\pm$ 0.025} & 1.293 $\pm$ 0.116 \\
PIDM  & \underline{0.107 $\pm$ 0.012} & 0.831 $\pm$ 0.061 \\
\textbf{PILD} (Ours)   & \textbf{0.074 $\pm$ 0.011} & \textbf{0.451 $\pm$ 0.055} \\

 \bottomrule
\end{tabular}
\caption{Quantitative results on plasma dynamics prediction.}
\label{table:plasma}
}

\end{center}
\end{table}

where $\mathcal{S}_n$ and $\mathcal{S}_{T_e}$ denote the right-hand-side source terms derived from the full two-fluid plasma model, which include effects such as advection, parallel heat conduction, and collisional coupling, and $\phi$ is electrostatic potential.

In our experiment, PILD is trained to jointly predict the fields of $n$ and $T_e$ based on observations including $\phi$ and other parameters.
We use the dataset from \citealp{mathews2021uncovering}, and compare our method against \textbf{PINN} \cite{mathews2021uncovering}, \textbf{DDPM} \cite{ho2020denoising}, \textbf{FNO} \cite{li2020fourier}, \textbf{DiffusionPDE} \cite{huang2024diffusionpde}, \textbf{PBFM} \cite{baldan2025flow}, and \textbf{PIDM} \cite{bastek2024physics}.

Quantitative results are reported in Table \ref{table:plasma}.
A representative sample is visualized in Figure \ref{fig:plasma0}, and more samples are provided in Figure \ref{fig:plasma} in Appendix \ref{plasma}.
The results indicate that incorporating physical information is beneficial for this complex PDE prediction problem under the matched experimental setup.
DDPM, which lacks explicit physics integration, exhibits larger errors overall.
PINN struggles to balance the physics loss with data distribution in generative tasks, leading to inferior performance compared to diffusion-based methods. 
The physics conditional alignment, Laplace-distributed residual modeling, and Jensen gap-aware scheduling collectively enable PILD to outperform methods such as DiffusionPDE and PIDM, achieving superior results across key metrics.

\subsection{Ablation Studies}

To disentangle methodological gains from architectural gains, all core ablations are conducted under the same task-specific DiT backbone, with identical training and sampling settings unless otherwise stated. 
We study the effects of the following components: (i) the probabilistic residual formulation, by comparing Gaussian and Laplace residual likelihoods; (ii) the Jensen-gap-aware adaptive residual scale; (iii) the physics-conditional alignment regularizer and (iv) the total number of diffusion timesteps. Results are in Table \ref{table:ablation_main}.

\begin{table}[t!]
\begin{center}
\setlength{\tabcolsep}{5pt}
\footnotesize
{
\begin{tabular}{l|c|c}
\toprule
\multirow{2}*{\textbf{Method}} & \multicolumn{1}{|c|}{\textbf{Plasma Dynamics}} & \multicolumn{1}{c}{\textbf{Tire Force}} \\ 
~ & Density error $\downarrow$ & $e_{F}\downarrow$ \\ 
\midrule
DiT (DDPM only) & 0.122 & 734.007 \\
+ Gaussian residual & 0.102 & 723.252 \\
+ Laplace, w/o Jensen & 0.088 & 713.044 \\
+ Laplace, w/ Jensen & 0.081 & 702.430 \\
+ Alignment alone & 0.083 & 719.795 \\
Full model & 0.074 & 695.494 \\
\midrule
Diffusion steps = 20 & 0.095 & 820.793 \\
Diffusion steps = 50 & 0.089 & 702.538 \\
Diffusion steps = 100 & 0.074 & 695.494 \\
\bottomrule
\end{tabular}
\caption{Core ablations of the proposed PILD framework under a shared DiT backbone. For Tire Force, the reported error is averaged over the three operating conditions.}
\label{table:ablation_main}
}
\end{center}
\end{table}

The first four rows evaluate the contribution of the probabilistic residual objective. Compared with the diffusion-only baseline, both Gaussian and Laplace residual modeling improve performance, while Laplace residuals consistently yield better results. Adding the Jensen-gap-aware correction further improves both benchmarks, confirming that the gain does not come solely from introducing a physics term, but also from correcting the bias induced by residual likelihood marginalization under noisy latent states.

The next two rows examine the effect of alignment. Adding the alignment term alone does not consistently improve performance, whereas the full model achieves the best results on both tasks. This indicates that the alignment mechanism is most effective when combined with the proposed probabilistic residual formulation and Jensen-gap-aware scaling.

Finally, increasing the total number of diffusion timesteps improves accuracy on both tasks. Here, the reported diffusion timesteps refer to the full discretization horizon of the forward/reverse diffusion process, rather than the number of DDIM reverse steps actually used for clean-state estimation or final generation.

Additional ablations on preprocessing, training cost, and other experimental details are provided in Appendix \ref{ablation}.

\section{Conclusion}
\label{conclusion}

This paper presents PILD, a physics-informed diffusion framework for improving the physical fidelity of generative models. By introducing a virtual residual observation with Laplace residual modeling and a Jensen-gap-aware adaptive residual scale, PILD incorporates physical constraints into diffusion training under a unified probabilistic objective. For conditional tasks, we further introduce a physics-conditional alignment mechanism to enhance consistency between latent representations and observation conditions. Experiments across engineering and scientific benchmarks demonstrate the effectiveness of the proposed framework.

A remaining limitation is that the Jensen gap is only mitigated rather than eliminated in a fully rigorous sense. Developing more principled approaches to remove this gap without introducing additional post-training or distillation stages remains an important direction for future work.

\section*{Acknowledgments}

The authors have stated that they have no potential conflicts of interest of this article and they did not receive any funding for this work.

\section*{Impact Statement}

The aim of this paper is to promote communication and research in the field of machine learning and to promote its application in cross-disciplines such as engineering and scientific fields. We do not think the possible social implications of this study need to be particularly emphasized here.


\bibliography{example_paper}
\bibliographystyle{icml2026}

\newpage
\appendix
\onecolumn
\section{Appendix}

\subsection{Preliminary}
\label{sup:pre}

\subsubsection{Diffusion Models}
\label{pre_diffusion}

Diffusion models learn to gradually convert a sample of a simple prior to sample from an unknown data distribution $q(\boldsymbol{x}_0)$ \cite{song2020denoising}. 
By adding Gaussian noise to sample data $\boldsymbol{x}_0 \sim q(\boldsymbol{x}_0)$, a forward process is defined as:
\begin{equation}
\begin{aligned}
    & q(\boldsymbol{x}_{1:T} |\boldsymbol{x}_0)=\prod^{T}_{t=1} q(\boldsymbol{x}_t | \boldsymbol{x}_{t-1}),\\
   & q(\boldsymbol{x}_t | \boldsymbol{x}_{t-1})=\mathcal{N}(\boldsymbol{x}_t ; \sqrt{\alpha_t} \boldsymbol{x}_{t-1},(1-\alpha _{t})\boldsymbol{I}),
\end{aligned}
\end{equation}
where $T$ is the number of forward steps and $\alpha_{t} \in(0,1)$ is a parameter related to step $t$. 
So $\boldsymbol{x}_t$ is expressed as a linear combination of $\boldsymbol{x}_0$ and a noise variable $\epsilon$:
\begin{equation}
\boldsymbol{x}_t=\sqrt{\alpha_t}\boldsymbol{x}_0+\sqrt{1-\alpha_t}\epsilon.
\end{equation}

A reverse process is defined to generate new samples:
\begin{equation}
\begin{aligned}
    &q(\boldsymbol{x}_{0:T})=p(\boldsymbol{x}_T) \prod ^{T}_{t=1} q(\boldsymbol{x}_{t-1} | \boldsymbol{x}_{t}),\\
    &q(\boldsymbol{x}_{t-1}|\boldsymbol{x}_t) = \mathcal{N}(\boldsymbol{x}_{t-1}; \boldsymbol{\mu}(\boldsymbol{x}_{t},t),\boldsymbol{\Sigma}(\boldsymbol{x}_{t},t)).
\end{aligned}
\end{equation}

We use $p_{\theta}(\boldsymbol{x}_0)$ generated by a neural network to approximate the distribution $q(\boldsymbol{x}_{0})$, which is defined as:
\begin{equation}
    p_\theta(\boldsymbol{x}_0)=\int p_\theta(\boldsymbol{x}_{0:T})\text{d}\boldsymbol{x}_{1:T},
\end{equation}
where $p_\theta(\boldsymbol{x}_{0:T}) := p_{\theta}(\boldsymbol{x}_T)\prod^{T}_{t=1}p_\theta^{(t)}(\boldsymbol{x}_{t-1}|\boldsymbol{x}_t)$ and $\boldsymbol{x}_1$, $\boldsymbol{x}_2$, $\cdots$, $\boldsymbol{x}_T$ are latent variables in the same sample space as $\boldsymbol{x}_0$.
The network aims to estimate the mean value $\boldsymbol{\mu}_\theta$,
while the covariance is fixed to:
\begin{equation}
\label{eq:10}
\boldsymbol{\Sigma}(\boldsymbol{x}_t,t)=\frac{1-\overline{\alpha}_{t-1}}{1-\overline{\alpha}_t}\beta_t \boldsymbol{I} = B_t\boldsymbol{I},
\end{equation}
where $\overline{\alpha}_t=\prod^{t}_{i=1}\alpha_i$, and $\beta_t=1-\alpha_t$.

The parameters $\theta$ are trained to fit $q(\boldsymbol{x}_0)$ by maximizing a variational lower bound \cite{song2020denoising}:
\begin{equation}
\label{eq:11}
\begin{aligned}
    &\max_\theta \mathbb{E}_{q(\boldsymbol{x}_0)}[\log p_{\theta}(\boldsymbol{x}_0)] \leq \\
    &\max_{\theta} \mathbb{E}_{q(\boldsymbol{x}_0,\boldsymbol{x}_1,\cdots,\boldsymbol{x}_{T})}[\log p_{\theta}(\boldsymbol{x_0};T)-\log q(\boldsymbol{x}_{1:T}|\boldsymbol{x}_{0})],
\end{aligned}
\end{equation}
where $q(\boldsymbol{x}_{1:T}|\boldsymbol{x}_{0})$ is inference distribution over latent variables $\boldsymbol{x}_1, \boldsymbol{x}_2, \cdots, \boldsymbol{x}_T$. 
Estimating $\boldsymbol{\mu}_\theta$ is obvious, but alternative parameterizations are possible.
According to the Bayes' theorem \cite{efron2013bayes}, the mean value $\boldsymbol{\mu}_t$ at time step $t$ is then given by:
\begin{equation}
\boldsymbol{\mu}_t(\boldsymbol{x}_t,\boldsymbol{\epsilon}_t)=\frac{\sqrt{\alpha_t}(1-\overline{\alpha}_{t-1})}{1-\overline{\alpha}_t}\boldsymbol{x}_t+\frac{\sqrt{\overline{\alpha}_{t-1}}\beta_t}{1-\overline{\alpha}_t}\boldsymbol{x}_0,
\end{equation}
where $\boldsymbol{\epsilon}_t$ is the applied Gaussian noise.
During training, $\boldsymbol{x}_t$ is known, so $\boldsymbol{\epsilon}_t$ is obtainable by predicting $\boldsymbol{\mu}_\theta$. 
Furthermore, \citealp{ho2020denoising} simplified the training loss as: 
\begin{equation}
\label{eq:13}
    \mathcal{L}_{\text{n}}(\theta):=\mathbb{E}_{t,\boldsymbol{x}_{0},\boldsymbol{\epsilon}}[||\boldsymbol{\epsilon}-\boldsymbol{\epsilon}_\theta (\sqrt{\overline{\alpha}_t}\boldsymbol{x}_0+\sqrt{1-\overline{\alpha}_t}\boldsymbol{\epsilon},t) ||^2],
\end{equation}
where $t$ is uniform from $1$ to $T$, $\boldsymbol{\epsilon}_\theta$ is the noise predicted by network with parameters $\theta$.
So, we can train the model to generate the clean signal $\boldsymbol{x}_0$.

\subsubsection{Physics-Informed Neural Network}
\label{app:pinn}

Typically, physical laws are expressed as a set of ODEs or PDEs.
For brevity, ODEs are regarded herein as PDEs with only one independent variable, and no additional preliminary of PINNs on ODEs will be provided.
The PDEs over a domain $\Omega \subset\mathbb{R}^d$ \cite{Raissi2019} is expressed as:
\begin{equation}
\begin{aligned}
\label{e1}
    \boldsymbol{\mathcal{D}}[\boldsymbol{u}(\boldsymbol{x}),\boldsymbol{x}] = 0, \text{ for }\boldsymbol{x} = (x_1,x_2,\cdots,x_d)^\top \in \Omega, 
\end{aligned}
\end{equation}
with boundary conditions:
\begin{equation}
     \boldsymbol{\mathcal{B}}[\boldsymbol{u}(\boldsymbol{x}),\boldsymbol{x}] = 0,\text{ for } \boldsymbol{x}  \in \partial\Omega, 
\end{equation}
where $\boldsymbol{\mathcal{D}}$ is the differential operator and $\boldsymbol{\mathcal{B}}$ defines the boundary conditions, $\partial\Omega$ is the boundary of $\Omega$, and $\boldsymbol{u}$ is modeled as the solution field for all data $\boldsymbol{x} \in \Omega$ satisfying the PDEs' set and boundary constraints.

In PINNs, the solution $\boldsymbol{u}(\boldsymbol{x})$ is approximated by a deep neural network, denoted as $\boldsymbol{\hat{u}}(\boldsymbol{x}; \boldsymbol{\theta})$, where $\boldsymbol{\theta}$ represents the set of all learnable parameters.

The discrepancy between the results calculated by the neural network and by physical model is represented by a residual $\boldsymbol{\mathcal{R}}(\boldsymbol{x}; \boldsymbol{\theta})$, which is defined by:
\begin{equation}
\label{e2}
    \boldsymbol{\mathcal{R}}(\boldsymbol{x}; \boldsymbol{\theta}):=\boldsymbol{\mathcal{D}}[\boldsymbol{\hat{u}}(\boldsymbol{x}; \boldsymbol{\theta}), \boldsymbol{x}],
\end{equation}
which also satisfies the boundary $\boldsymbol{\mathcal{B}}[\boldsymbol{\hat{u}}(\boldsymbol{x}; \boldsymbol{\theta}), \boldsymbol{x}]=0$.

The parameters $\boldsymbol{\theta}$ are optimized by minimizing a physical loss $\mathcal{L}_{\text{p}}(\boldsymbol{\theta})$, which typically includes a term for the PDE residual and another term for the boundary deviation penalty:
\begin{equation}
\begin{aligned}
\label{e3}
    \mathcal{L}_{\text{p}}(\boldsymbol{\theta}) =  
    \frac{1}{N_f} \sum_{i=1}^{N_f} \| \boldsymbol{\mathcal{R}}(\boldsymbol{x}_f^i; \boldsymbol{\theta}) \|_2^2 + \frac{1}{N_b} \sum_{i=1}^{N_b} \| \boldsymbol{\mathcal{B}}[\boldsymbol{\hat{u}}(\boldsymbol{x}_b^i; \boldsymbol{\theta}), \boldsymbol{x}_b^i] \|_2^2,
\end{aligned}
\end{equation}
where $\{ \boldsymbol{x}_f^i \}_{i=1}^{N_f}$ is a set of $N_f$ collocation points sampled from the domain $\Omega$, where the PDE residual is enforced, and $\{ \boldsymbol{x}_b^i \}_{i=1}^{N_b}$ is a set of $N_b$ points sampled from the boundary $\partial\Omega$, where the boundary conditions are enforced.

A Data-driven loss $\mathcal{L}_{\text{d}}(\theta)$ is presented as:
\begin{equation}
\label{e6}
    \mathcal{L}_{\text{d}}(\boldsymbol{\theta}) = \frac{1}{N_d} \sum_{i=1}^{N_d} \| \boldsymbol{\hat{u}}(\boldsymbol{x}_d^i; \boldsymbol{\theta}) - \boldsymbol{u}^i \|_2^2,
\end{equation}
where $\boldsymbol{u}^i$ here is the data ground truth.
Total loss $\mathcal{L}(\boldsymbol{\theta})$ becomes weighted sum of $\mathcal{L}_{\text{p}}(\boldsymbol{\theta})$ and $\mathcal{L}_{\text{d}}(\boldsymbol{\theta})$.

\subsection{Consistency and Jensen-Gap Correction}
\label{A1}

To theoretically validate the consistency of the proposed PILD framework, we analyze the model from the perspective of score-based generative modeling \cite{song2020score}, extending the continuous-time diffusion formulation to incorporate physical constraints. This section shows that incorporating the physics-informed residual term preserves distributional consistency, while also clarifying the two approximation gaps discussed in the main text: the posterior-representative gap and the additional Jensen-type bias induced by residual likelihood marginalization.

\subsubsection{Continuous-Time Diffusion Formulation}

We formalize the forward diffusion process as a stochastic differential equation (SDE) over $t \in [0,T]$:
\begin{equation}
\mathrm{d}\boldsymbol{x}_t = \boldsymbol{F}_t \boldsymbol{x}_t \mathrm{d}t + \boldsymbol{G}_t \mathrm{d}\boldsymbol{w}_t,
\end{equation}
where $\boldsymbol{w}_t$ denotes a standard Wiener process, $\boldsymbol{F}_t \in \mathbb{R}^{D \times D}$ is the drift coefficient, and $\boldsymbol{G}_t \in \mathbb{R}^{D \times D}$ is the diffusion coefficient. For consistency with the discrete diffusion setup in the main text, we adopt the parameterization aligned with DDPM \cite{ho2020denoising}:
\begin{equation}
\boldsymbol{F}_t = \frac{1}{2} \frac{\mathrm{d}\log \overline{\alpha}_t}{\mathrm{d}t} \boldsymbol{I}, 
\qquad 
\boldsymbol{G}_t = \sqrt{-\frac{\mathrm{d}\log \overline{\alpha}_t}{\mathrm{d}t}} \boldsymbol{I},
\end{equation}
where $\overline{\alpha}_t$ decreases from $\overline{\alpha}_0 \approx 1$ to $\overline{\alpha}_T \approx 0$.

The reverse diffusion process follows \cite{zhang2022fast}:
\begin{equation}
\label{eq:SDE}
\mathrm{d}\boldsymbol{x}_t = \left[ \boldsymbol{F}_t \boldsymbol{x}_t - \frac{1+\lambda^2}{2} \boldsymbol{G}_t \boldsymbol{G}_t^\top \nabla_{\boldsymbol{x}_t} \log p_t(\boldsymbol{x}_t) \right] \mathrm{d}\overline{t} + \lambda \boldsymbol{G}_t \mathrm{d}\overline{\boldsymbol{w}}_t,
\end{equation}
where $\overline{t}=T-t$ is the reversed time, $\overline{\boldsymbol{w}}_t$ is a reversed Wiener process, and $\lambda \ge 0$ controls the stochasticity.

\subsubsection{Physics-Informed Score Matching Objective}

The extended score-matching objective is
\begin{equation}
\label{eq:extended}
\begin{aligned}
\boldsymbol{s}_{\mathrm{opt}}
=
\arg\min_{\boldsymbol{s}}
\mathbb{E}_{t \sim \mathrm{Unif}[0,T]}
\mathbb{E}_{q(\boldsymbol{x}_0) q(\boldsymbol{x}_t|\boldsymbol{x}_0)}
\Big[
&\Lambda(t) \left\| \nabla_{\boldsymbol{x}_t} \log q(\boldsymbol{x}_t|\boldsymbol{x}_0) - \boldsymbol{s}(\boldsymbol{x}_t,t,\boldsymbol{O}) \right\|_2^2 \\
&\quad - \log q_{\mathcal{R}}(\hat{\boldsymbol{r}} = \boldsymbol{0} \mid \boldsymbol{x}_0^*(\boldsymbol{x}_t,t,\boldsymbol{O}))
\Big],
\end{aligned}
\end{equation}
where $\Lambda(t)>0$ is a time-dependent weighting factor, $\boldsymbol{x}_0^*$ is the denoised estimate induced by the reverse process, and
\begin{equation}
q_{\mathcal{R}}(\hat{\boldsymbol{r}} \mid \boldsymbol{x}_0^*)
=
\mathrm{Laplace}(\hat{\boldsymbol{r}};\mathcal{R}(\boldsymbol{x}_0^*), b_t \boldsymbol{I}).
\end{equation}

We now state the following proposition:

Let $q(\boldsymbol{x}_0)$ be a data distribution whose samples satisfy the physical constraint $\mathcal{R}(\boldsymbol{x}_0)=\boldsymbol{0}$. If the optimal score $\boldsymbol{s}_{\mathrm{opt}}$ minimizes the extended objective in Equation \eqref{eq:extended}, then solving the reverse SDE \eqref{eq:SDE} with $\boldsymbol{s}_{\mathrm{opt}}$ generates samples from the true distribution $\boldsymbol{x}_0 \sim q(\boldsymbol{x}_0)$.

\begin{proof}
For standard diffusion models, perfect score recovery $\boldsymbol{s}(\boldsymbol{x}_t,t)=\nabla_{\boldsymbol{x}_t}\log p_t(\boldsymbol{x}_t)$ ensures that the reverse SDE's marginal distribution matches the forward diffusion distribution for all $t$ \cite{zhang2022fast}. Sampling from $p(\boldsymbol{x}_T)\sim \mathcal{N}(\boldsymbol{0},\boldsymbol{I})$ and solving the reverse SDE thus recovers $\boldsymbol{x}_0 \sim q(\boldsymbol{x}_0)$.

For the physics-informed term, under the Laplace model we have
\begin{equation}
-\log q_{\mathcal{R}}(\hat{\boldsymbol{r}}=\boldsymbol{0}\mid \boldsymbol{x}_0^*)
=
\frac{1}{b_t}\|\mathcal{R}(\boldsymbol{x}_0^*)\|_1 + C,
\end{equation}
where $C$ is constant with respect to $\boldsymbol{x}_0^*$ and $\boldsymbol{s}$. Since $\boldsymbol{x}_0 \sim q(\boldsymbol{x}_0)$ satisfies $\mathcal{R}(\boldsymbol{x}_0)=\boldsymbol{0}$, the optimal score model that matches the true data score is also compatible with minimizing the residual term. Hence the extended objective does not alter the target distribution but only penalizes deviations from its physical manifold. Therefore, solving the reverse SDE with $\boldsymbol{s}_{\mathrm{opt}}$ generates samples from $q(\boldsymbol{x}_0)$ that satisfy the physical constraints.
\end{proof}

\subsubsection{Approximation Gaps from Residual Evaluation and Marginalization}

Beyond the usual posterior-mean mismatch
\begin{equation}
\mathcal{R}\!\left(\mathbb{E}[\boldsymbol{x}_0 \mid \boldsymbol{x}_t,\boldsymbol{O}]\right)
\neq
\mathbb{E}\!\left[\mathcal{R}(\boldsymbol{x}_0) \mid \boldsymbol{x}_t,\boldsymbol{O}\right],
\end{equation}
there exists a second source of bias due to the residual likelihood marginalization itself. In the main text, the first source is mitigated by replacing the conditional mean with a DDIM-based estimator $\boldsymbol{x}_0^*$, which is closer to evaluating the residual on actual samples, although it remains an approximation to the full posterior clean-state distribution.

To analyze the second source, let
\begin{equation}
S_t := \|\mathcal{R}(\boldsymbol{x}_0)\|_1,
\qquad
\boldsymbol{x}_0 \sim p_\theta(\boldsymbol{x}_0 \mid \boldsymbol{x}_t,\boldsymbol{O}),
\end{equation}
and define
\begin{equation}
\mu_t := \mathbb{E}[S_t \mid \boldsymbol{x}_t,\boldsymbol{O}],
\qquad
\sigma_t^2 := \mathrm{Var}(S_t \mid \boldsymbol{x}_t,\boldsymbol{O}).
\end{equation}
Under the Laplace virtual likelihood,
\begin{equation}
q_{\mathcal{R}}(\hat{\boldsymbol{r}}=\boldsymbol{0}\mid \boldsymbol{x}_0)
\propto
\exp\left(-\frac{S_t}{b_t}\right),
\end{equation}
so the exact negative physical log-likelihood is
\begin{equation}
\label{eq:true_free_energy_appendix}
\mathcal{L}^{\mathrm{true}}_{\mathrm{phys}}(t)
=
-\log
\mathbb{E}
\left[
\exp\left(-\frac{S_t}{b_t}\right)
\,\middle|\,
\boldsymbol{x}_t,\boldsymbol{O}
\right].
\end{equation}

Setting $Z_t := -S_t/b_t$ and using the cumulant expansion of $\log \mathbb{E}[e^{Z_t}]$, we obtain
\begin{equation}
\log \mathbb{E}[e^{Z_t}]
=
\mathbb{E}[Z_t]
+
\frac{1}{2}\mathrm{Var}(Z_t)
+
\frac{1}{6}\kappa_{3,t}
+\cdots,
\end{equation}
where $\kappa_{3,t}$ is the third conditional cumulant of $Z_t$. Substituting $Z_t=-S_t/b_t$ yields
\begin{equation}
\label{eq:free_energy_expand_appendix}
\mathcal{L}^{\mathrm{true}}_{\mathrm{phys}}(t)
=
\frac{\mu_t}{b_t}
-
\frac{\sigma_t^2}{2b_t^2}
+
\frac{1}{6b_t^3}\kappa_3(S_t \mid \boldsymbol{x}_t,\boldsymbol{O})
+\mathcal{O}(b_t^{-4}).
\end{equation}
Hence, to second order,
\begin{equation}
\label{eq:second_order_true_phys_appendix}
\mathcal{L}^{\mathrm{true}}_{\mathrm{phys}}(t)
\approx
\frac{\mu_t}{b_t}
-
\frac{\sigma_t^2}{2b_t^2}.
\end{equation}

This shows that the log-marginalization bias is governed by the conditional residual variance $\sigma_t^2$, in addition to the diffusion-induced base scale $b_t$.

To absorb this correction into a stable positive residual-penalty form, we introduce the effective scale $\tilde b_t$ by matching
\begin{equation}
\frac{\mu_t}{\tilde b_t}
\approx
\frac{\mu_t}{b_t}
-
\frac{\sigma_t^2}{2b_t^2}.
\end{equation}
Rearranging gives
\begin{equation}
\tilde b_t
\approx
\frac{b_t}{1-\frac{\sigma_t^2}{2\mu_t b_t}}.
\end{equation}
Since Equation \eqref{eq:second_order_true_phys_appendix} is itself only a second-order approximation, we do not retain this rational form in practice. Instead, applying the first-order expansion
\begin{equation}
(1-z)^{-1} \approx 1+z,
\qquad
z=\frac{\sigma_t^2}{2\mu_t b_t},
\end{equation}
motivates the practical stabilized effective scale
\begin{equation}
\label{eq:effective_bt_appendix}
\tilde b_t
=
b_t
+
\frac{\sigma_t^2}{2\mu_t+\varepsilon},
\end{equation}
with $\varepsilon>0$ a small stability constant. 
This should be understood as a practical first-order stabilization rather than an exact algebraic identity. 
Moreover, the diffusion-induced base scale $b_t$ itself provides an intrinsic stabilizing effect: since it lower-bounds the effective residual scale, it reduces the relative impact of noise in the variance-based correction term, especially at high-noise timesteps.

Its use is further supported by the training behavior discussed in Appendix \ref{A4}: as the KL divergence between $q(\bm{x}_{1:T})$ and $p_\theta(\bm{x}_{1:T})$ rapidly decreases during optimization, the approximation of the residual expectation under $p_\theta$ by its counterpart under $q$ becomes increasingly accurate in practice. 
Accordingly, this construction yields the practical correction used in the main text: DDIM-based residual evaluation mitigates the posterior-representative gap, while $\tilde b_t$ provides a variance-aware correction for the leading-order bias induced by residual likelihood log-marginalization.

In practice, raw timestep-wise minibatch plug-in estimates of the residual moments can still be noisy, especially in early training. To improve stability, we smooth the minibatch estimates using exponential moving averages (EMA). Specifically, for timestep $t$ at training iteration $k$, let $\hat{\mu}_t^{(k)}$ and $\hat{\sigma}_t^{2,(k)}$ denote the minibatch estimates from Equation \eqref{eq:mu_sigma_est}. We maintain the EMA statistics
\begin{equation}
\label{eq:ema_mu_appendix}
\bar{\mu}_t^{(k)}
=
\rho \,\bar{\mu}_t^{(k-1)}
+
(1-\rho)\,\hat{\mu}_t^{(k)},
\end{equation}
\begin{equation}
\label{eq:ema_sigma_appendix}
\bar{\sigma}_t^{2,(k)}
=
\rho \,\bar{\sigma}_t^{2,(k-1)}
+
(1-\rho)\,\hat{\sigma}_t^{2,(k)},
\end{equation}
where $\rho \in [0,1)$ is the EMA momentum. The practical effective scale used in training is then given by
\begin{equation}
\label{eq:effective_bt_ema_appendix}
\hat{\tilde b}_t^{(k)}
=
b_t
+
\frac{\bar{\sigma}_t^{2,(k)}}{2(\bar{\mu}_t^{(k)}+\varepsilon)}.
\end{equation}
Therefore, the effective residual scale remains anchored to the stable base scale $b_t$ while the variance-based correction is smoothed over iterations, reducing sensitivity to noisy minibatch fluctuations.

\subsection{Laplace Distribution}
\label{A2}

In our framework, the virtual residual observation is modeled using a Laplace distribution. This choice is theoretically and practically motivated for several reasons.

First, compared to the Gaussian distribution, the Laplace distribution exhibits heavier tails, making it more robust to outliers and large deviations. The probability density function of a univariate Laplace distribution centered at zero is
\begin{equation}
p(r \mid b) = \frac{1}{2b} \exp\left(-\frac{|r|}{b}\right),
\end{equation}
where $r$ denotes the residual and $b > 0$ is the scale parameter. In contrast, the Gaussian density decays quadratically in the exponent, leading to stronger sensitivity to large residuals.

Second, from an optimization perspective, the Laplace residual model corresponds to an $\ell_1$-type physical penalty:
\begin{equation}
-\log q_{\mathcal{R}}(\hat{\boldsymbol{r}}=\boldsymbol{0} \mid \boldsymbol{x})
\propto
\frac{1}{b}\|\mathcal{R}(\boldsymbol{x})\|_1.
\end{equation}
This promotes robustness to noisy or imperfect data while still encouraging adherence to the physical constraint manifold.

Third, the Laplace likelihood is log-concave and yields well-behaved gradients, which improves optimization stability in practice. Empirically, we observe that it performs better than Gaussian residual modeling, especially on noisy real-world engineering datasets.

We also emphasize that PILD does not require the data distribution to satisfy the governing physics exactly. Rather, our assumption is that the data conform to the correct physical principles in a probabilistic sense. The Laplace residual model is specifically introduced to accommodate realistic noise, bias, and moderate deviations while still enforcing physically meaningful structure.

In summary, the Laplace distribution provides a favorable trade-off between fidelity to physical laws and robustness to imperfect observations, making it well suited for modeling virtual residual observations.

\subsection{Unified Training Objective}
\label{A4}

The original optimization objective involves sampling latent trajectories from the learned distribution $\bm{x}_{1:T} \sim p_\theta(\bm{x}_{1:T})$. To make this computationally tractable, we instead sample from the known forward process $q(\bm{x}_{1:T})$. This approximation effectively disregards the likelihood ratio term and introduces an additional bias, which is expected to diminish as optimizing the variational lower bound drives $p_\theta(\bm{x}_{1:T})$ toward $q(\bm{x}_{1:T})$:
\begin{equation}
\label{eq33}
\begin{aligned}
&\mathbb{E}_{\bm{x}_0 \sim q(\bm{x}_0)} \left[ -\log p_\theta(\bm{x}_0) \right]
+ \mathbb{E}_{\bm{x}_{1:T} \sim p_\theta(\bm{x}_{1:T})}
\left[ -\log q_{\mathcal{R}}(\hat{\bm{r}} = \bm{0} \mid \bm{x}_0^*(\bm{x}_{1:T})) \right] \\
&\quad \leq
\mathbb{E}_{q(\bm{x}_{0:T})}
\left[
\log \frac{q(\bm{x}_{1:T} \mid \bm{x}_0)}{p_\theta(\bm{x}_{0:T})}
\right]
+
\mathbb{E}_{\bm{x}_{1:T} \sim p_\theta(\bm{x}_{1:T})}
\left[
-\log q_{\mathcal{R}}(\hat{\bm{r}} = \bm{0} \mid \bm{x}_0^*(\bm{x}_{1:T}))
\right] \\
&\quad \approx
\mathbb{E}_{q(\bm{x}_{0:T})}
\left[
\log \frac{q(\bm{x}_{1:T} \mid \bm{x}_0)}{p_\theta(\bm{x}_{0:T})}
-
\log q_{\mathcal{R}}(\hat{\bm{r}} = \bm{0} \mid \bm{x}_0^*(\bm{x}_{1:T}))
\right].
\end{aligned}
\end{equation}

Building on Equation \eqref{eq33}, the data-driven component remains the standard diffusion loss
\begin{equation}
\lambda_t \|\epsilon - \epsilon_\theta\|^2,
\end{equation}
while the physics-informed component is replaced by the effective-scale form
\begin{equation}
\frac{1}{\hat{\tilde b}_t}
\left\|
\mathcal{R}(\bm{x}_0^*)
\right\|_1,
\qquad
\hat{\tilde b}_t
=
b_t
+
\frac{\hat{\sigma}_t^2}{2(\hat{\mu}_t+\varepsilon)}.
\end{equation}
Here $b_t = B_t/c$ is the diffusion-induced base residual scale, while $\hat{\mu}_t$ and $\hat{\sigma}_t^2$ estimate the conditional mean and variance of the residual magnitude.

By unifying these two components under the same expectation over \(t \sim [1, T]\), \(x_0 \sim q(x_0)\), \(O\), and \(\epsilon \sim \mathcal{N}(0, I)\), we obtain the base loss
\begin{equation}
\begin{aligned}
\label{eq:final_appendix_loss}
\mathcal{L}_{\text{base}}(\boldsymbol{\theta})
=
\mathbb{E}_{t \sim [1,T], (\boldsymbol{x}_0, \boldsymbol{O}) \sim q(\boldsymbol{x}_0, \boldsymbol{O}), \boldsymbol{\epsilon} \sim \mathcal{N}(\mathbf{0}, \mathbf{I})}
\Big[
\lambda_t \| \boldsymbol{\epsilon} - \boldsymbol{\epsilon}_\theta \|^2 
+
\frac{1}{\hat{\tilde b}_t}
\left\|
\boldsymbol{\mathcal{R}}({\boldsymbol{x}}_0^*)
\right\|_1
\Big].
\end{aligned}
\end{equation}
This objective preserves the original diffusion training structure while replacing heuristic residual weighting by a principled leading-order correction derived from the log-marginalized residual likelihood.

To further support the approximation $q(\bm{x}_{1:T}) \approx p_\theta(\bm{x}_{1:T})$ during training, we calculate the KL divergence between $q$ and $p_{\theta}$, with results shown in Table \ref{table:app_KL}.

\begin{table}[htbp!]
\begin{center}
\setlength{\tabcolsep}{15pt}
\footnotesize
{
\begin{tabular}{ll}
\toprule
\textbf{Iter} & \textbf{KL divergence} \\
\midrule
$1$ & $5.210e+01$ \\
$0.2$k & $2.615e-02$ \\
$2$k & $3.446e-03$ \\
$20$k & $1.661e-03$ \\
$100$k & $3.103e-04$ \\
$200$k & $2.648e-04$ \\
$300$k & $3.982e-04$ \\
\bottomrule
\end{tabular}
\caption{KL divergence between $q$ and $p_{\theta}$.}
\label{table:app_KL}
}
\end{center}
\end{table}

The KL divergence drops by over three orders of magnitude within the first 2k iterations. The minor fluctuation at 300k iterations reflects stochastic optimization noise rather than divergence, as the value remains within the same order of magnitude. These results provide empirical support for the approximation $q(\bm{x}_{1:T}) \approx p_\theta(\bm{x}_{1:T})$ during training, suggesting that the likelihood-ratio mismatch becomes small throughout the vast majority of optimization. While this does not constitute a formal error bound for the expectation replacement in Equation \eqref{eq33}, it supports the practical adequacy of the approximation used to derive the unified training objective.

\subsection{Details of Network Architecture}
\label{A5}

The model training is conducted on an RTX 4090/24GB GPU. 
The setup of each experiment differs slightly, and the detailed configurations are provided in Tables \ref{table:appendix1} to \ref{table:appendix3}. 
Each benchmark is trained independently with a task-specific DiT backbone, so the hidden dimension, Transformer depth, patch size, and MAE encoder design are allowed to vary across tasks.

\begin{table}[htbp!]
\begin{center}
\setlength{\tabcolsep}{15pt}
\footnotesize

{
\begin{tabular}{ll}
\toprule
\multirow{1}*{\textbf{Hyperparameter}} & \multicolumn{1}{l}{\textbf{Value}} \\
\midrule
Backbone & 1D DiT \\
Patch size & $4$ \\
Hidden dimension & $256$ \\
Transformer blocks & $6$ \\
Attention heads & $8$ \\
MLP ratio & $4$ \\
Time embedding dimension & $128$ \\
Observation embedding dimension & $128$ \\
Alignment layer $l$ & $4$ \\
Projection head & 2-layer MLP \\
MAE encoder & 1D MAE \\
MAE patch size & $4$ \\
MAE encoder dimension & $128$ \\
MAE encoder blocks & $4$ \\
MAE decoder blocks & $2$ \\
MAE mask ratio & $0.5$ \\
Diffusion timesteps & $100$ \\
Batch size & $8$ \\
Iterations & $10^4$ \\
Initial learning rate & $10^{-3}$ \\
Optimization algorithm & Adam \cite{adam2014method} \\
Physical penalty weight $c$ & $0.005$ \\
Alignment weight $\lambda_{\mathrm{align}}$ & $0.01$ \\
$\beta_t$-scheduler & Cosine schedule \cite{dhariwal2021diffusion} \\
$\beta_1$,$\beta_t$ & $10^{-4}, 0.03$ \\
\bottomrule
\end{tabular}
\caption{Implementation details of tracking task.}
\label{table:appendix1}
}

\end{center}
\end{table}

\begin{table}[htbp!]
\begin{center}
\setlength{\tabcolsep}{15pt}
\footnotesize

{
\begin{tabular}{ll}
\toprule
\multirow{1}*{\textbf{Hyperparameter}} & \multicolumn{1}{l}{\textbf{Value}} \\
\midrule
Backbone & 1D DiT \\
Patch size & $8$ \\
Hidden dimension & $384$ \\
Transformer blocks & $10$ \\
Attention heads & $12$ \\
MLP ratio & $4$ \\
Time embedding dimension & $128$ \\
Observation embedding dimension & $128$ \\
Alignment layer $l$ & $6$ \\
Projection head & 2-layer MLP \\
MAE encoder & 1D MAE \\
MAE patch size & $8$ \\
MAE encoder dimension & $192$ \\
MAE encoder blocks & $6$ \\
MAE decoder blocks & $2$ \\
MAE mask ratio & $0.5$ \\
Diffusion timesteps & $100$ \\
Batch size & $32$ \\
Iterations & $5 \times 10^5$ \\
Initial learning rate & $10^{-5}$ \\
Optimization algorithm & Adam \cite{adam2014method} \\
Physical penalty weight $c$ & $0.0001$ \\
Alignment weight $\lambda_{\mathrm{align}}$ & $0.01$ \\
$\beta_t$-scheduler & Cosine schedule \cite{dhariwal2021diffusion} \\
$\beta_1$,$\beta_t$ & $10^{-4}, 0.03$ \\
\bottomrule
\end{tabular}
\caption{Implementation details of tire force estimation.}
\label{table:appendix4}
}

\end{center}
\end{table}

\begin{table}[htbp!]
\begin{center}
\setlength{\tabcolsep}{15pt}
\footnotesize

{
\begin{tabular}{ll}
\toprule
\multirow{1}*{\textbf{Hyperparameter}} & \multicolumn{1}{l}{\textbf{Value}} \\
\midrule
Backbone & 2D DiT \\
Patch size & $4 \times 4$ \\
Hidden dimension & $256$ \\
Transformer blocks & $8$ \\
Attention heads & $8$ \\
MLP ratio & $4$ \\
Time embedding dimension & $128$ \\
Observation embedding dimension & - \\
Alignment layers $\mathcal{S}$ & - \\
Projection head & - \\
MAE encoder & - \\
MAE patch size & - \\
MAE encoder dimension & - \\
MAE encoder blocks & - \\
MAE decoder blocks & - \\
MAE mask ratio & - \\
Diffusion timesteps & $100$ \\
Batch size & $8$ \\
Iterations & $3 \times 10^5$ \\
Initial learning rate & $10^{-4}$ \\
Optimization algorithm & Adam \cite{adam2014method} \\
Physical penalty weight $c$ & $0.001$ \\
Alignment weight $\lambda_{\mathrm{align}}$ & - \\
$\beta_t$-scheduler & Cosine schedule \cite{dhariwal2021diffusion} \\
$\beta_1$,$\beta_t$ & $10^{-6}, 10^{-2}$ \\
\bottomrule
\end{tabular}
\caption{Implementation details of Darcy flow.}
\label{table:appendix2}
}

\end{center}
\end{table}

\begin{table}[htbp!]
\begin{center}
\setlength{\tabcolsep}{15pt}
\footnotesize

{
\begin{tabular}{ll}
\toprule
\multirow{1}*{\textbf{Hyperparameter}} & \multicolumn{1}{l}{\textbf{Value}} \\
\midrule
Backbone & 2D DiT \\
Patch size & $4 \times 4$ \\
Hidden dimension & $512$ \\
Transformer blocks & $12$ \\
Attention heads & $16$ \\
MLP ratio & $4$ \\
Time embedding dimension & $128$ \\
Observation embedding dimension & $128$ \\
Alignment layer $l$ & $8$ \\
Projection head & 2-layer MLP \\
MAE encoder & 2D MAE \\
MAE patch size & $8 \times 8$ \\
MAE encoder dimension & $256$ \\
MAE encoder blocks & $6$ \\
MAE decoder blocks & $2$ \\
MAE mask ratio & $0.75$ \\
Diffusion timesteps & $100$ \\
Batch size & $32$ \\
Iterations & $5 \times 10^5$ \\
Initial learning rate & $10^{-5}$ \\
Optimization algorithm & Adam \cite{adam2014method} \\
Physical penalty weight $c$ & $0.00001$ \\
Alignment weight $\lambda_{\mathrm{align}}$ & $0.01$ \\
$\beta_t$-scheduler & Cosine schedule \cite{dhariwal2021diffusion} \\
$\beta_1$,$\beta_t$ & $10^{-6}, 10^{-2}$ \\
\bottomrule
\end{tabular}
\caption{Implementation details of plasma dynamics.}
\label{table:appendix3}
}

\end{center}
\end{table}

We emphasize that the reported ``Diffusion timesteps = 100'' in the above tables refers to the total discretization horizon of the diffusion process, i.e., the number of timesteps used to define the forward noising process and its corresponding reverse-time parameterization. It does not mean that 100 reverse denoising steps are executed whenever a clean-state representative is needed. In our implementation, practical clean-state estimation from an intermediate noisy state is carried out with a reduced two-step DDIM procedure. Specifically, during training, for a sampled noisy state at timestep $t$, we use a two-step DDIM-based estimator to obtain $\boldsymbol{x}_0^*$ for residual evaluation; during inference, we likewise use a two-step DDIM sampler for final generation. Therefore, the tabulated diffusion timestep number specifies the underlying diffusion horizon, whereas the actual DDIM reverse steps used in practice are reduced to two.

For conditional tasks, the MAE encoder is pretrained in a separate self-supervised stage using only the corresponding observation $\boldsymbol{O}$. For 1D tasks, we use a 1D patch-based MAE that masks temporal patches and reconstructs the missing segments. For 2D tasks, we use a standard patch-based MAE that masks spatial patches and reconstructs the missing regions. After pretraining, the MAE decoder is discarded and only the encoder is retained. Its parameters are then frozen during diffusion training and used to provide the target representation for the alignment loss. 

In our implementation, alignment is applied only at a selected intermediate DiT block rather than at all layers. The layer index is specified by $\mathcal{S}$ in Tables \ref{table:appendix1}--\ref{table:appendix3}.
This choice provides a compact condition-consistency signal while avoiding the additional cost of aligning every block. Freezing the MAE encoder further prevents the alignment target from drifting together with the diffusion backbone and stabilizes the representation regularization. To further reduce late-stage optimization instability, we additionally employ early stopping based on the validation loss.

It should be noted that the tracking, tire-force, and plasma tasks are treated as conditional generation problems and therefore use both observation embeddings and the alignment regularizer. By contrast, Darcy flow is modeled as an unconditional generation task, so neither observation conditioning nor MAE-based alignment is applied in that case.


\subsection{Inference and Sampling Details}
\label{A5_sampling}

For inference, we follow \cite{bastek2024physics} and use a two-step DDIM sampler for efficiency. The full diffusion process is defined over $T$ timesteps, but final generation uses only a reduced two-step reverse trajectory. Starting from $\boldsymbol{x}_T \sim \mathcal{N}(\mathbf{0}, \mathbf{I})$, we first map $\boldsymbol{x}_T$ to an intermediate state $\boldsymbol{x}_1$ using the DDIM update, and then reconstruct the final clean sample $\boldsymbol{x}_0$ from $\boldsymbol{x}_1$.

\begin{algorithm}[h!]
\caption{Two-Step DDIM Sampling for PILD}
\label{alg:sampling}
\begin{algorithmic}[1]
\STATE {\bfseries Input}: Trained noise prediction network $\boldsymbol{\epsilon}_\theta$, condition $\boldsymbol{O}$, total diffusion timesteps $T$, cumulative variance schedule $\overline{\alpha}_t$
\STATE {\bfseries Output}: Generated sample $\boldsymbol{x}_0$

\STATE Draw $\boldsymbol{x}_T \sim \mathcal{N}(\mathbf{0}, \mathbf{I})$
\STATE Predict noise at timestep $T$:
\[
\hat{\boldsymbol{\epsilon}} = \boldsymbol{\epsilon}_\theta(\boldsymbol{x}_T, \boldsymbol{O}, T)
\]
\STATE Estimate the clean state from $\boldsymbol{x}_T$:
\[
\hat{\boldsymbol{x}}_0^{(T)} =
\frac{1}{\sqrt{\overline{\alpha}_T}}
\left(
\boldsymbol{x}_T
-
\sqrt{1-\overline{\alpha}_T}\,\hat{\boldsymbol{\epsilon}}
\right)
\]
\STATE Compute the intermediate state $\boldsymbol{x}_1$ via the DDIM update:
\[
\boldsymbol{x}_1
=
\sqrt{\overline{\alpha}_1}\,\hat{\boldsymbol{x}}_0^{(T)}
+
\sqrt{\frac{1-\overline{\alpha}_1}{1-\overline{\alpha}_T}}
\left(
\boldsymbol{x}_T
-
\sqrt{\overline{\alpha}_T}\,\hat{\boldsymbol{x}}_0^{(T)}
\right)
\]
\STATE Predict noise at timestep $1$:
\[
\hat{\boldsymbol{\epsilon}}' = \boldsymbol{\epsilon}_\theta(\boldsymbol{x}_1, \boldsymbol{O}, 1)
\]
\STATE Reconstruct the final sample:
\[
\boldsymbol{x}_0
=
\frac{1}{\sqrt{\overline{\alpha}_1}}
\left(
\boldsymbol{x}_1
-
\sqrt{1-\overline{\alpha}_1}\,\hat{\boldsymbol{\epsilon}}'
\right)
\]
\end{algorithmic}
\end{algorithm}

This sampling procedure should be distinguished from the total number of diffusion timesteps used to define the underlying diffusion process. In our experiments, the diffusion horizon is set to $T=100$, whereas actual inference uses only the reduced two-step DDIM trajectory above.

\subsection{Other Formations of Constraints}
\label{app:otherform}

The proposed PILD framework is not restricted to PDE/ODE residuals and can be extended to algebraic equations and inequality constraints. 
Below we show that these constraint types can be incorporated into the same Jensen-gap-aware training form.

\subsubsection{Algebraic Equation Constraints}

Algebraic equations describe relationships between variables without involving derivatives, e.g., equilibrium conditions in structural design, material property constraints, or algebraic identities in physical systems. 
Consider an algebraic constraint of the form
\begin{equation}
\mathcal{A}(x_0) = a_{\text{target}}, \quad x_0 \sim p_\theta(x_0 | O),
\end{equation}
where $\mathcal{A}: \mathbb{R}^D \to \mathbb{R}^k$ denotes the algebraic operator and $a_{\text{target}} \in \mathbb{R}^k$ is the target value. 
We define the corresponding residual as
\begin{equation}
\mathcal{R}_{\text{alg}}(x_0^*) = \mathcal{A}(x_0^*) - a_{\text{target}},
\end{equation}
where $x_0^* = \mathrm{DDIM}[x_t, O, t]$ is the denoised estimate. 
Following the main framework, we model the virtual residual observation with a Laplace likelihood:
\begin{equation}
q_{\mathcal{R}_{\text{alg}}}(\hat{r}_{\text{alg}} | x_0^*) = {Laplace}\left(\hat{r}_{\text{alg}}; \mathcal{R}_{\text{alg}}(x_0^*), b_t I\right).
\end{equation}
The resulting training loss is
\begin{equation}
\mathcal{L}_{\text{PILD-alg}}(\theta)
=
\mathbb{E}_{t \sim [1,T], (\boldsymbol{x}_0, \boldsymbol{O}) \sim q(\boldsymbol{x}_0, \boldsymbol{O}), \boldsymbol{\epsilon} \sim \mathcal{N}(\mathbf{0}, \mathbf{I})}
\left[
\lambda_t \left\| \epsilon - \epsilon_\theta \right\|^2
+
\frac{1}{\hat{\tilde b}_t}
\left\| \mathcal{R}_{\text{alg}}(x_0^*) \right\|_1
\right].
\end{equation}

\subsubsection{Inequality Constraints}

Inequality constraints are ubiquitous in practical problems. 
Consider a general inequality constraint
\begin{equation}
\mathcal{H}(x_0) \leq h_{\text{max}}, \quad x_0 \sim p_\theta(x_0 | O),
\end{equation}
where $\mathcal{H}: \mathbb{R}^D \to \mathbb{R}^m$ is the inequality operator and $h_{\text{max}} \in \mathbb{R}^m$ is the upper bound.
We transform this inequality into a residual using the ReLU operator \cite{bastek2024physics}:
\begin{equation}
\label{relu}
\mathcal{R}_{\text{ineq}}(x_0^*) = \mathrm{ReLU}\left(\mathcal{H}(x_0^*) - h_{\text{max}}\right).
\end{equation}
This residual vanishes when the inequality is satisfied and is positive otherwise.
We model it as a Laplace virtual observation:
\begin{equation}
q_{\mathcal{R}_{\text{ineq}}}(\hat{r}_{\text{ineq}} | x_0^*) = {Laplace}\left(\hat{r}_{\text{ineq}}; \mathcal{R}_{\text{ineq}}(x_0^*), b_t I\right).
\end{equation}
The corresponding training loss is
\begin{equation}
\mathcal{L}_{\text{PILD-ineq}}(\theta)
=
\mathbb{E}_{t \sim [1,T], (\boldsymbol{x}_0, \boldsymbol{O}) \sim q(\boldsymbol{x}_0, \boldsymbol{O}), \boldsymbol{\epsilon} \sim \mathcal{N}(\mathbf{0}, \mathbf{I})}
\left[
\lambda_t \left\| \epsilon - \epsilon_\theta \right\|^2
+
\frac{1}{\hat{\tilde b}_t}
\left\| \mathcal{R}_{\text{ineq}}(x_0^*) \right\|_1
\right].
\end{equation}

\subsubsection{Toy Experiment}

\begin{figure}[h!]
    \centering
    \includegraphics[width=0.95\linewidth]{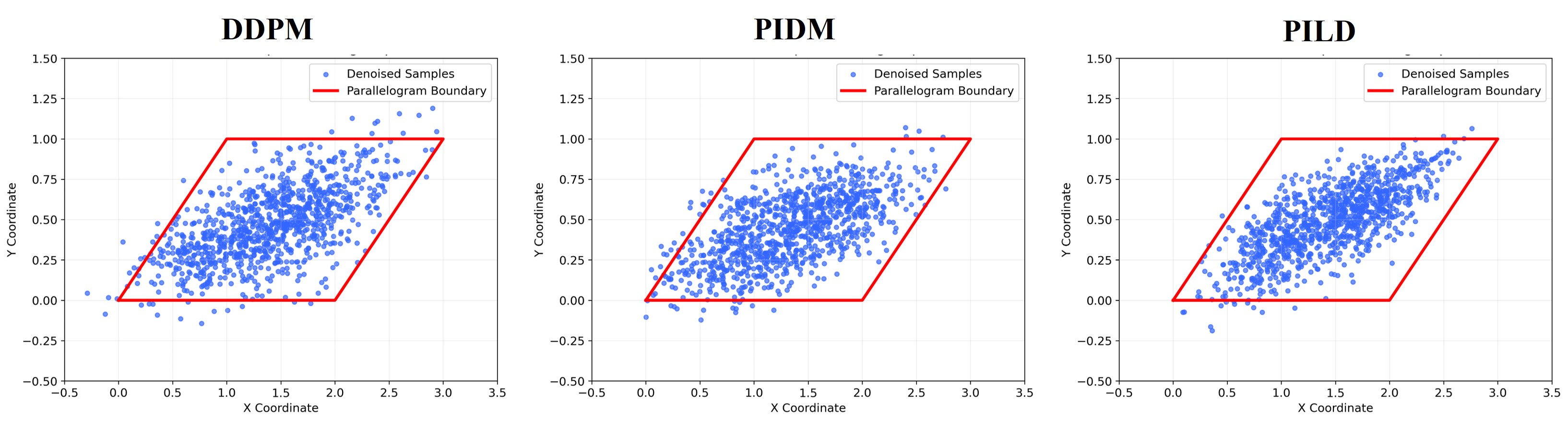}
    \caption{Results of toy experiment. The red parallelogram indicates the feasible region, and the blue points denote sampled data.}
    \label{fig:toy}
\end{figure}

Since both equality and inequality constraints can be cast into the same residual form, we present a toy experiment for the more general inequality setting. 
We define four inequalities that form a parallelogram-shaped feasible region and sample 10,000 training points uniformly from this region. 
The constraint in Equation \eqref{relu} is then incorporated into the PILD objective. 
The results of DDPM, PIDM, and PILD are shown in Figure \ref{fig:toy}.

All three methods perform reasonably well on this simple problem. 
However, DDPM, which only learns the data distribution, produces more outliers near the sharp boundaries. 
Both physics-informed methods improve boundary adherence, and PILD produces slightly fewer outliers than PIDM, which is consistent with the more robust residual modeling used in our framework.

\subsection{Details of Experimental Setup}
\label{A6}

\subsubsection{Description of Other Comparative Methods}
\label{othermethods}
As our four benchmarks span two distinct domains with different data modalities, the selection of domain-appropriate baselines is necessary.

The first two experimental scenarios are engineering application scenarios, where the comparative methods include common methods for the corresponding scenarios and methods from the same broad category as PILD. Specifically, EKF is a traditional method widely used for state estimation in nonlinear engineering systems; FCN and ResNet are classic neural-network predictors that directly learn the mapping between input and output data; the two PINN-based methods incorporate physical loss terms into the corresponding neural architectures, which is a commonly adopted practice in physics-informed neural networks for engineering systems; B-PINN introduces Bayesian operations to handle system noise; the two LSTM-based methods are designed for time-series data; and DDPM and PIDM provide diffusion-based references. These baselines cover heterogeneous modeling paradigms, so we use them to evaluate practical task performance rather than to claim that all model classes have identical objectives. For fairness, all methods are evaluated using the same input variables, preprocessing steps, train/test splits, normalization factors, and metrics within each benchmark. Open-source implementations are used when available for LSTM-RNN, B-PINN, DDPM, and PIDM, while the remaining methods are reproduced based on their original papers and tuned under the same validation protocol.

The latter two experiments correspond to scientific machine learning scenarios, where the comparative methods include both common operator-learning baselines and representative physics-informed generative models. Specifically, DDPM serves as the standard diffusion baseline (DiT) trained only on data likelihood and therefore does not explicitly enforce physical constraints during training. FNO is a representative operator-learning model, providing a strong non-generative reference for PDE surrogate modeling. PINN is included as a classical physics-informed baseline that enforces governing equations through residual regularization, but it is not designed to model complex generative distributions. CoCoGen is a score-based generative approach that improves physical consistency mainly through inference-time correction under PDE guidance, rather than modifying the training objective itself. DiffusionPDE learns a joint generative prior over coefficient and solution spaces and uses PDE-guided sampling under partial observation; however, its physical constraints are primarily imposed during guided inference instead of being incorporated into a unified residual likelihood during training. PIDM is the most closely related prior work on physics-informed diffusion training: it augments the diffusion objective with a Gaussian residual penalty evaluated on a DDIM-based or mean-based clean-state representative, using a fixed or timestep-coupled scale. PBFM is a closely related physics-constrained generative baseline based on flow matching rather than diffusion, and combines residual minimization with conflict-free gradient updates and unrolling to alleviate the representative mismatch during training.

Our framework is motivated by a different design objective: rather than appending a generic physics penalty to diffusion training, we aim to make the physical component better aligned with realistic residual statistics and better integrated with the conditional generation process. First, many practical engineering and scientific datasets contain measurement noise, modeling mismatch, and occasional outliers, so we model the virtual residual observation with a Laplace likelihood, which provides a more robust probabilistic interpretation of residual minimization than a Gaussian alternative. Second, once physical residuals are introduced into diffusion training, the residual likelihood is evaluated under noisy latent states, which gives rise to a Jensen-type marginalization bias. To account for this effect, we introduce a Jensen-gap-aware effective residual scale derived from the log-marginalized residual likelihood, so that the physical penalty is modulated in a statistically motivated manner rather than by a purely fixed or heuristic weighting rule. Third, because our framework is built on a DiT backbone for conditional tasks, we further introduce a physics-conditional alignment mechanism to encourage intermediate latent representations to remain consistent with the observation conditions throughout denoising. Taken together, these components are intended to provide a more principled formulation of physics-informed generative learning, with emphasis on physically meaningful residual modeling and stronger condition-aware physical guidance.

Regarding architectural differences, PIDM and DiffusionPDE are originally implemented with U-Net-style backbones in their released or reported settings, whereas our method uses a DiT backbone. Therefore, the benchmark comparisons in the main experiments should be interpreted as system-level comparisons under matched data, preprocessing, and evaluation protocols, rather than as pure backbone-controlled comparisons. 
To avoid conflating methodological gains with architectural changes, all core ablations in our paper are conducted under the same DiT backbone, where we separately evaluate the effects of Laplace residual modeling, the Jensen-gap-aware residual scale, and the alignment mechanism. 
It should be noted that the DiT backbone combined with Gaussian modeling in the ablation setting can be regarded as the DiT-based PIDM. However, our ablation studies show that even when PIDM is equipped with a DiT backbone, it still underperforms compared to PILD, thereby demonstrating the effectiveness of our proposed design.
To ensure the accuracy of the cited referance, we retain the original open-source implementation of PIDM in our main experiments.
In addition, whenever preprocessing is applied, the same preprocessing pipeline is used for all compared methods within each benchmark, and all baselines are trained or fine-tuned under the same train/test splits, input variables, normalization factors, and evaluation metrics. For methods with released pretrained weights, we fine-tune them under our evaluation protocol to avoid unfair under-training; for methods without compatible checkpoints, we reproduce them as faithfully as possible following their original papers.

\subsubsection{Ablation on denoising preprocessing}
\label{ablation}
It should be noted that we evaluate the performance of different methods on real-world datasets with noise after applying denoising as a preprocessing step using Gaussian filtering \cite{ito2002gaussian} in the main text. 
Here, the experiment demonstrates the performance of various methods on experimental inputs without any denoising applied.
We conduct the experiment on the Tire Force dataset, and the results are reported in Table \ref{table:ablation2}. 
It can be observed that denoising improves the performance of all methods to some extent. 
However, since denoising may suppress fine-grained features in the data, other approaches struggle to effectively reconcile physical models with the underlying data distribution. 
In contrast, PILD maintains competitive performance, demonstrating its robustness and ability to jointly leverage physics and data characteristics.

\begin{table}[htbp!]

\begin{center}
\setlength{\tabcolsep}{5pt} 
\footnotesize

{
\begin{tabular}{l|c|c}
\toprule
\multirow{1}*{\textbf{Method}} & Denoised Input & Original Input \\  
\midrule

PINN-ResNet & 774.011 & 957.683\\
DDPM  & 763.612 & 781.054\\
PIDM  &732.910 & 828.933 \\
\textbf{PILD}(Ours)   &695.494 & 775.914 \\

 \bottomrule
\end{tabular}
\caption{Ablation studies on input denoising by Gaussian filtering.}
\label{table:ablation2}
}

\end{center}
\end{table}

\begin{figure}[htpb!]
    \centering
    \includegraphics[width=1\linewidth]{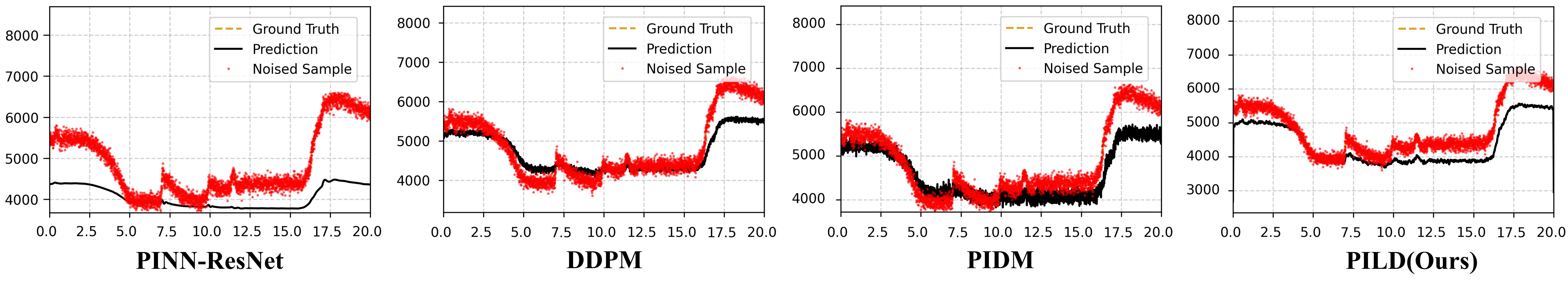}
    \caption{Ablation results with noisy input on Tire Force benchmark.}
    \label{fig:ablation noise}
\end{figure}

We also select one group of samples generated with original noisy input without any preprocessing in Figure \ref{fig:ablation noise}.

With the original noisy input, PINN-Res deviates noticeably from the target, and PIDM does not fully remove the noise in this sample. 
In this sample, both DDPM and PILD achieve satisfactory performance: PILD denoises the data more thoroughly but introduces a larger deviation than DDPM. 
However, in terms of the overall test results, PILD achieves a slightly lower error. The presentation of this sample also illustrates the importance of data preprocessing.

\subsubsection{Training cost}
We provide a comparison of run time and memory cost among representative baseline methods in Table \ref{table:app_runtime}. The reported training time refers only to the main diffusion-model training stage.

\begin{table}[ht!]
\begin{center}
\setlength{\tabcolsep}{10pt}
\footnotesize
{
\begin{tabular}{l|l|l|l|l}
\toprule
\multirow{1}*{\textbf{Condition}} & \multirow{1}*{\textbf{Method}} & \textbf{Training time} & \textbf{Inf. time} & \textbf{Memory} \\  
\midrule

\multirow{3}*{\textbf{Tracking}} 
& PINN & 19min & 0.1s & 1.5G \\

~ & DDPM & 22min & 4.5s & 2.1G \\

~ & PILD & 35min & 0.08s & 2.6G \\

\midrule

\multirow{4}*{\textbf{Darcy}} 
& DDPM & 5.5h & 10s & 3.1G \\

~ & PBFM & 8.5h & 0.16s & 3.5G \\

~ & PIDM & 8h & 0.19s & 3.2G \\

~ & PILD & 8h & 0.18s & 3.2G \\

\bottomrule
\end{tabular}
\caption{Comparison of run time and memory cost. The reported training time includes only the diffusion-model training stage.}
\label{table:app_runtime}
}
\end{center}
\end{table}

PILD requires moderately higher training cost than standard diffusion baselines because it incorporates additional physics-informed residual evaluation and alignment regularization during optimization. This overhead mainly affects training, while inference remains nearly unchanged due to the use of the same reduced two-step DDIM procedure. For unconditional tasks such as Darcy flow, the training cost of PILD remains comparable to other physics-informed diffusion baselines.

\subsubsection{Vehicle Tracking}
\label{app:tracking}
\begin{figure}[htbp!]
    \centering
    \includegraphics[width=0.5\linewidth]{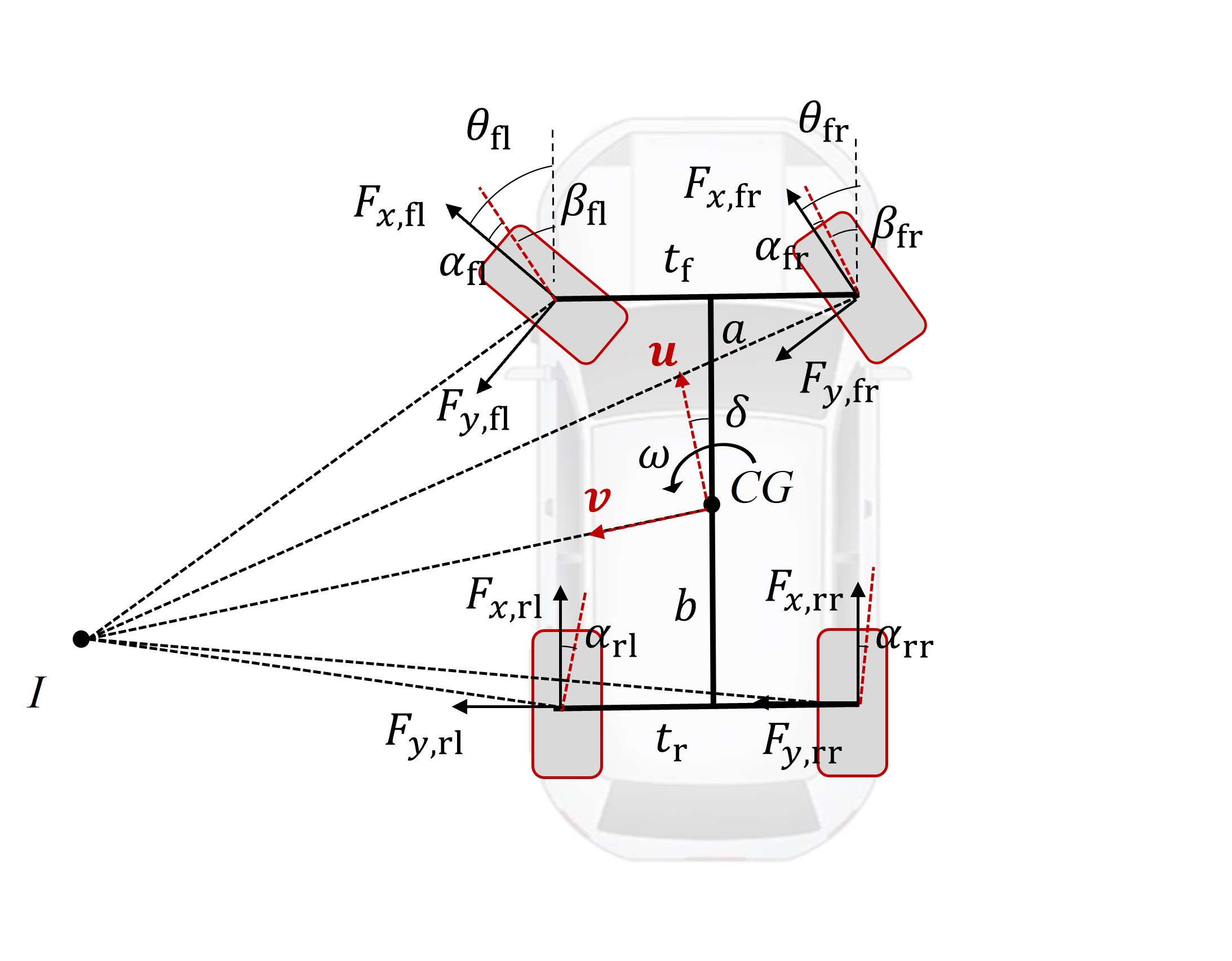}
    \caption{Ackerman model for tracking.}
    \label{fig:ackerman}
\end{figure}

When a vehicle is steering, the angles of the left and right wheels differ to enable both wheels to travel around the steering center simultaneously. 
This constitutes the most important feature of the Ackerman model, as illustrated in Figure \ref{fig:ackerman}. 
Furthermore, since the wheels are made of rubber, the lateral force during vehicle steering originates from the elastic force generated by rubber deformation. 
Consequently, there is an angle between the direction of the wheel's traveling speed and the geometric orientation of the wheel, which is known as the slip angle. 
The direction of the wheel's traveling speed is indicated by the red dashed line in Figure \ref{fig:ackerman}.
In Figure \ref{fig:ackerman}, $I$ is the vehicle's instantaneous center of steering; $CG$ is the vehicle's center of mass; $u$, \(v\), and \(\omega\) are the longitudinal velocity, lateral velocity, and yaw rate at the vehicle's center of mass position, respectively, and the direction of the lateral velocity \(v\) points to the instantaneous center; the angle \(\delta\) between the longitudinal velocity \(u\) and the vehicle's longitudinal axis is the side slip angle of the center of mass. 
\(t_{\text{f}}\), and \(t_{\text{r}}\) are the front and rear track widths, respectively; \(a\) and \(b\) are the distances from the center of mass to the front and rear axles.

Based on the above analysis, we can derive the following relationships:
\begin{equation}
\begin{aligned}
\label{eq:x}
m(\dot{u} - v\omega) =\sum F_u = & F_{x,\mathrm{fr}}\cos(\theta_{\mathrm{fr}} + \delta) - F_{y,\mathrm{fr}}\sin(\theta_{\mathrm{fr}} + \delta) + 
F_{x,\mathrm{fl}}\cos(\theta_{\mathrm{fl}} + \delta) - F_{y,\mathrm{fl}}\sin(\theta_{\mathrm{fl}} + \delta) + \\ &  (F_{x,\mathrm{rr}} + F_{x,\mathrm{rl}})\cos\delta - (F_{y,\mathrm{rr}} + F_{y,\mathrm{rl}})\sin\delta,\\
\end{aligned}
\end{equation}

\begin{equation}
\begin{split}
\label{eq:y}
m(\dot{v} - u\omega) =\sum F_v =& F_{x,\mathrm{fr}}\sin(\theta_{\mathrm{fr}} + \delta) + F_{y,\mathrm{fr}}\cos(\theta_{\mathrm{fr}} + \delta) + 
F_{x,\mathrm{fl}}\sin(\theta_{\mathrm{fl}} + \delta) + F_{y,\mathrm{fl}}\cos(\theta_{\mathrm{fl}} + \delta) + \\
&(F_{x,\mathrm{rr}} + F_{x,\mathrm{rl}})\sin\delta + (F_{y,\mathrm{rr}} + F_{y,\mathrm{rl}})\cos\delta,\\
\end{split}
\end{equation}
\begin{equation}
\begin{split}
\label{eq:z}
&I_z \dot{\omega} = \sum M_z = F_{x,\text{fr}} d_1 + F_{y,\text{fr}} d_2 - F_{x,\text{fl}} d_3 + F_{y,\text{fl}} d_4 + F_{x,\text{rr}} t_{\text{r}}/2 - F_{y,\text{rr}} b - F_{x,\text{rl}} t_{\text{r}}/2 - F_{y,\text{rl}} b = 0, \\
\end{split}
\end{equation}
\begin{equation}
\theta_j = \alpha_j + \beta_j.
\label{eq:theta}
\end{equation}

In Equations \eqref{eq:x} to \eqref{eq:theta}, \(F_u\) and \(F_v\) are the components of the adhesive force of each tire in the directions of the longitudinal velocity \(u\) and the lateral velocity \(v\), respectively; \(M_z\) is the yaw moment of the adhesive force of each tire about the center of mass \(O\); \(m\) is the vehicle mass; \(I_z\) is the moment of inertia of the vehicle about the vertical axis passing through the center of mass; \(F_{xi}\) is the longitudinal adhesive force of the tire, \(i = \text{fr}, \text{fl}, \text{rr}, \text{rl}\); \(F_{yi}\) is the lateral adhesive force of the tire; \(\theta_j\) is the front wheel steering angle, \(j = \text{fr}, \text{fl}\); \(\alpha_i\) is the tire side slip angle; \(\beta_i\) is the angle between the tire velocity direction and the vehicle's longitudinal axis:
\begin{equation}
\begin{split}
    &\beta_{\text{fr}} = \arctan (\frac{a-R \sin \delta}{R \cos \delta+t_\text{f}/2}), \quad
    \beta_{\text{fl}} = \arctan (\frac{a-R \sin \delta}{R \cos \delta-t_\text{f}/2}), \quad \\
    & \beta_{\text{fr}}=\alpha_{\text{rr}} = \arctan (\frac{b+R \sin \delta}{R \cos \delta+t_\text{f}/2}), \quad
    \beta_{\text{fr}}=\alpha_{\text{rl}} = \arctan (\frac{b+R \sin \delta}{R \cos \delta-t_\text{f}/2}).
    \end{split}
\end{equation}

\(d_1\), \(d_2\), \(d_3\), and \(d_4\) are the force arms of the tire adhesive force about the vehicle's center of mass, respectively:
\begin{equation}
\begin{split}
    d_1=\sqrt{(t_\text{f}/2)^2+a^2}\sin{(\arctan \frac{t_f}{2a}+(\beta_{\text{fr}}+\alpha_{\text{fr}}))}, \quad
    d_2=\sqrt{(t_\text{f}/2)^2+a^2}\cos{(\arctan \frac{t_f}{2a}+(\beta_{\text{fr}}+\alpha_{\text{fr}}))}, \\
    d_3=\sqrt{(t_\text{f}/2)^2+a^2}\sin{(\arctan \frac{t_f}{2a}-(\beta_{\text{fr}}+\alpha_{\text{fr}}))}, \quad
    d_4=\sqrt{(t_\text{f}/2)^2+a^2}\cos{(\arctan \frac{t_f}{2a}-(\beta_{\text{fr}}+\alpha_{\text{fr}}))}.
    \end{split}
\end{equation}

Once we have the above relationships, we define a state variable as:
\begin{equation}
    \boldsymbol{x}=[x,y,\psi,v]\top,
\end{equation}
where, 
\begin{equation}
    \dot{x} = v \cos(\psi),
\end{equation}
\begin{equation}
\dot{y} = v \sin(\psi),
\end{equation}
\begin{equation}
\begin{aligned}
\dot{\psi} = \omega = \frac{1}{I_z} \sum M_z =& \frac{1}{I_z} [ F_{x,\text{fr}} d_1 + F_{y,\text{fr}} d_2-F_{x,\text{fl}} d_3 + F_{y,\text{fl}} d_4 + F_{x,\text{rr}} \frac{t_{\text{r}}}{2} -F_{y,\text{rr}} b - F_{x,\text{rl}} \frac{t_{\text{r}}}{2} - F_{y,\text{rl}} b ],
\end{aligned}
\end{equation}
\begin{equation}
\dot{v} = \frac{1}{m} \left( \sum F_u \cos(\psi) + \sum F_v \sin(\psi) \right).
\end{equation}

It should be noted that the experimental setup is designed solely to validate the effectiveness of the proposed method; therefore, data from other perception sensors commonly used in autonomous driving are excluded from training. 
Furthermore, trajectory prediction here is not performed by generating an entire trajectory in one shot. Instead, the model predicts one future point at a time and uses that prediction as input for the next step. This setting is adopted to evaluate and compare the accumulation of prediction errors over time against other methods.

\begin{table}[htbp!]

\begin{center}
\setlength{\tabcolsep}{8pt} 

\footnotesize

\begin{tabular}{l|ccc|ccc}
\toprule
\multirow{2}*{\textbf{Method}} & \multicolumn{3}{|c|}{\textbf{Downtown Driving}} & \multicolumn{3}{|c}{\textbf{Rural Driving}} \\ 
~ & \textbf{$e_{x\&y}$$\downarrow$}          & \textbf{$e_\psi$$\downarrow$}          & \textbf{$e_{v_x}$$\downarrow$}    & \textbf{$e_{x\&y}$$\downarrow$}          & \textbf{$e_\psi$$\downarrow$}          & \textbf{$e_{v_x}$$\downarrow$}\\

\midrule

EKF &  6.602$\pm$1.980 &  6.930$\pm$1.705 & 5.476$\pm$1.628 &  6.179$\pm$1.372 & 6.241$\pm$1.524& 5.512$\pm$1.208 \\
FCN &  9.567$\pm$2.891 &  7.977$\pm$2.060 & 14.284$\pm$5.179 &  9.623$\pm$3.034 & 6.606$\pm$2.706& 10.972$\pm$3.167 \\

ResNet & 9.236$\pm$2.769 & 6.899$\pm$1.592 & 9.094$\pm$2.772 & 5.517$\pm$0.989 &10.361$\pm$2.974& 9.315$\pm$1.601 \\

PINN-FCN & 8.824$\pm$1.062 & 14.488$\pm$2.965& 17.287$\pm$3.907 &8.279$\pm$1.291 &10.349$\pm$1.352&15.251$\pm$3.740  \\

PINN-ResNet  &5.923$\pm$0.993 &9.753$\pm$1.367 & 11.286$\pm$2.073 & \underline{3.729$\pm$0.595} & 9.134$\pm$1.394 & 8.248$\pm$1.449\\

B-PINN  &\underline{5.867$\pm$0.936} &6.597$\pm$0.883 & 10.563$\pm$1.192 & 6.034$\pm$0.937 & 6.511$\pm$0.861 & 9.656$\pm$1.104 \\

LSTM-RNN & 8.953$\pm$1.392 & 10.199$\pm$1.495 & 11.298$\pm$2.968 & 9.643$\pm$1.420 &8.831$\pm$1.631& 8.168$\pm$1.093 \\

LSTM-GA & 9.019$\pm$1.022 & 6.579$\pm$0.846 & 8.442$\pm$0.781 & 9.005$\pm$1.923 & \underline{5.994$\pm$0.821} & 10.837$\pm$1.907 \\

DDPM & 9.653$\pm$0.911 & \underline{6.109$\pm$0.693} & \underline{4.317$\pm$0.404} & 4.697$\pm$0.453 &7.138$\pm$0.560& \underline{4.936$\pm$0.390}\\

\textbf{PILD} (Ours) & \textbf{4.796$\pm$0.501} &  \textbf{5.391$\pm$0.359} &  \textbf{3.048$\pm$0.296} & \textbf{3.562$\pm$0.304} &  \textbf{5.837$\pm$0.306} &  \textbf{3.033$\pm$0.289}  \\ \bottomrule

\end{tabular}
\caption{Quantitative results with sample-level standard deviations on tracking tasks. The $\pm$ values are computed across evaluated trajectories or generated samples in a single trained-model evaluation.}

\label{table:app:tracking}
\end{center}
\end{table}

To verify the generalization ability across different vehicle models, we trained the model using data from multiple vehicle types (i.e., vehicles with distinct parameters) simultaneously. Additionally, due to the significant noise present in the vehicle chassis data, we have performed denoising on the raw data. The necessity of this denoising process is also discussed in Appendix \ref{ablation}.
We performed statistical analysis to determine data magnitude and applied fixed scaling factors during computation, then reversed the scaling afterward to normalize the data.
The metrics of the tracking task are as follows: $e_{x\&y}$ is the deviation (m) to the target point; $e_{\psi}$ is the heading error (deg); and $e_{V_x}$ is the speed error (km/h).

\subsubsection{Tire Force}
\label{tireforce}

Tire dynamic load refers to the dynamic vertical force exerted on the tire during vehicle motion. Due to the complexity of the vehicle chassis structure (incorporating components such as springs and dampers), directly calculating this load based solely on vehicle body attitude and acceleration introduces significant errors. This approach fails to accurately reflect the actual force experienced by the tire, thereby limiting the ability to properly evaluate chassis performance.

\begin{figure}[htbp!]
    \centering
    \includegraphics[width=0.75\linewidth]{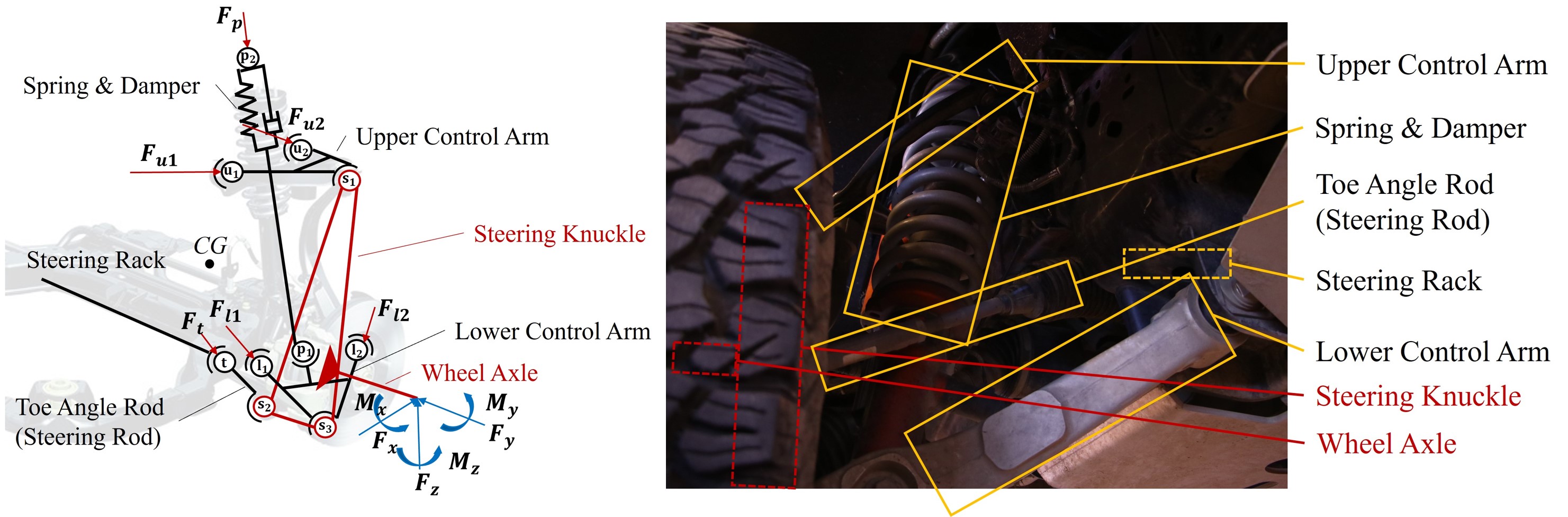}
    \caption{Physical suspension system.}
    \label{fig:suspension}
\end{figure}

During vehicle operation, the spring \& damper assembly undergoes expansion and contraction in response to relative displacement between the road surface and vehicle body. Since suspensions typically consist of multiple connecting rods forming complex single-degree-of-freedom (1DOF) or two-degree-of-freedom (2DOF) structures, suspension geometric parameters change dynamically during driving. A photograph of the actual suspension system is shown in Figure \ref{fig:suspension}.

Taking a double-wishbone suspension as an example, we simplify it to the linkage structure shown in Figure \ref{fig:suspension}, comprising rods, ball joints, a spring, and a damper. The red component in the figure represents the steering knuckle, which connects directly to the wheel. Blue forces and moments indicate forces transmitted from the wheel to the suspension, while red forces represent internal forces within suspension rods.

For computational simplicity, we neglect the mass of individual suspension links, considering only the masses of the wheels and steering knuckles. Suspension hinge points are referred to as hard points, whose coordinates (for those not fixed to the vehicle body) change during driving. We define the hard points as follows:
$\bold{u_1}$ and $\bold{u_2}$ (front and rear points of the upper control arm, UCA);
$\bold{l_1}$ and $\bold{l_2}$ (front and rear points of the lower control arm, LCA);
$\bold{p_1}$ and $\bold{p_2}$ (lower and upper joints of the spring-damper assembly);
$\bold{t}$ (hinge joint between the steering knuckle and steering rack);
$\bold{s_1}$, $\bold{s_2}$, and $\bold{s_3}$ (upper, front, and lower points of the steering knuckle).

Here, $CG$ denotes the center of mass; $\boldsymbol{F_p}$, $\boldsymbol{F_{u1}}$, $\boldsymbol{F_{u2}}$, $\boldsymbol{F_t}$, $\boldsymbol{F_{l1}}$, $\boldsymbol{F_{l2}}$ represent suspension force vectors; $\boldsymbol{F_x}$, $\boldsymbol{F_y}$, $\boldsymbol{F_z}$ are tire forces transmitted from the wheel; and $\boldsymbol{M_x}$, $\boldsymbol{M_y}$, $\boldsymbol{M_z}$ are moments about the wheel axle. 

We define the vector from $CG$ to hard point $i$ as $\overrightarrow{i}$ (e.g., $\overrightarrow{\bold{u_1}}$) and the direction vector from hard point $i$ to $j$ as $\overrightarrow{ij}$ (e.g., $\overrightarrow{\bold{u_1}\bold{s_1}}$). Force vectors can thus be expressed as:

\begin{equation}
\begin{aligned}
    &\boldsymbol{F}_p = F_p \cdot \overrightarrow{\bold{p_2}\bold{p_1}}, \quad 
    \boldsymbol{F}_{u1} = F_{u1} \cdot \overrightarrow{\bold{u_1}\bold{s_1}}, \quad
    \boldsymbol{F}_{u2} = F_{u2} \cdot \overrightarrow{\bold{u_2}\bold{s_1}}, \\  
    &\boldsymbol{F}_{t} = F_{t} \cdot \overrightarrow{\bold{t}\bold{s_2}}, \quad
    \boldsymbol{F}_{l1} = F_{l1} \cdot \overrightarrow{\bold{l_1}\bold{s_3}}, \quad 
    \boldsymbol{F}_{l2} = F_{l2} \cdot \overrightarrow{\bold{l2}\bold{s_3}},
\end{aligned}
\end{equation}

where unbolded coefficients represent scalar magnitudes of the corresponding forces. The suspension force matrix is:

\begin{equation}
    \boldsymbol{F_{\text{sus}}} = \left[ \boldsymbol{F}_p, \boldsymbol{F}_{u1}, \boldsymbol{F}_{u2}, \boldsymbol{F}_{t}, \boldsymbol{F}_{l1}, \boldsymbol{F}_{l2} \right]
\end{equation}

Moments of these forces about $CG$ are calculated as:

\begin{equation}
\begin{aligned}
    &\boldsymbol{M}_p = \overrightarrow{\bold{p_2}} \times \boldsymbol{F_p}, \quad 
    \boldsymbol{M}_{u1} = \overrightarrow{\bold{u_1}} \times \boldsymbol{F_{u1}}, \quad
    \boldsymbol{M}_{u2} = \overrightarrow{\bold{u_2}} \times \boldsymbol{F_{u2}}, \\
   & \boldsymbol{M}_{t} = \overrightarrow{\bold{t}} \times \boldsymbol{F_t}, \quad
    \boldsymbol{M}_{l1} = \overrightarrow{\bold{l_1}} \times \boldsymbol{F_{l1}}, \quad 
    \boldsymbol{M}_{l2} = \overrightarrow{\bold{l_2}} \times \boldsymbol{F_{l2}}.
\end{aligned}
\end{equation}

The moment matrix is:

\begin{equation}
    \boldsymbol{M_{\text{sus}}} = \left[ \boldsymbol{M}_p, \boldsymbol{M}_{u1}, \boldsymbol{M}_{u2}, \boldsymbol{M}_{t}, \boldsymbol{M}_{l1}, \boldsymbol{M}_{l2} \right].
\end{equation}

Force and moment equilibrium relationships yield:

\begin{equation}
\label{eq24}
    \sum_{i \in \{p,u1,u2,t,l1,l2\}} \boldsymbol{F}_i = \sum_{j \in \{x,y,z\}} \boldsymbol{F}_j, \quad
    \sum_{i \in \{p,u1,u2,t,l1,l2\}} \boldsymbol{M}_i = \sum_{j \in \{x,y,z\}} \boldsymbol{M}_j.
\end{equation}

As shown in Figure \ref{fig:suspension}, the front suspension has two degrees of freedom: steering and bump. We denote the steering rack displacement as $x_a$ and the spring-damper expansion/contraction as $x_d$, both measurable by sensors. These parameters directly influence suspension geometry, so hard point coordinates are expressed as functions of $x_a$ and $x_d$ (e.g., $\overrightarrow{\bold{u_1}(x_a,x_d)}$, $\overrightarrow{\bold{u_2}(x_a,x_d)}$).

With a displacement sensor installed on the spring-damper assembly, $\boldsymbol{F_p}$ can be determined as:

\begin{equation}
    \boldsymbol{F_p}(t) = m \cdot f\left(\frac{\partial^2 x_d(t)}{\partial t^2}\right) + c \cdot \frac{\partial x_d(t)}{\partial t} + k \cdot x_d(t),
\end{equation}

where $f\left(\frac{\partial^2 x_d(t)}{\partial t^2}\right)$ represents the nonlinear function of unsprung mass $m$ acceleration acting on the spring-damper assembly, accounting for inertial forces.

According to the tire model, both $\boldsymbol{F_x}$ and $\boldsymbol{F_y}$ can be expressed by $\boldsymbol{F_z}$ as:

The Pacejka Magic Formula provides a mathematically rigorous relationship between tire forces and operating conditions, expressed as:

\begin{equation}
\begin{aligned}
\label{e27}
    F_y(\alpha, F_z, T) = D(T) \sin[ C(T) \arctan(B(T) \alpha - E(T)(B(T) \alpha - \arctan(B(T) \alpha))) ], \\
    F_x(k, F_z, T) = D_x(T) \sin[ C_x(T) \arctan(B_x(T) s - E_x(T) (B_x(T) s - \arctan(B_x(T) s)))],
\end{aligned}
\end{equation}

where, $\alpha$ is the slip angle, $s$ is the longitudinal slip ratio; $B(T), C(T), D(T), E(T)$ are temperature-dependent coefficients for lateral forces; $B_x(T), C_x(T), D_x(T), E_x(T)$ are temperature-dependent coefficients for longitudinal forces; $T$ represents tire temperature, which significantly affects rubber compound properties.

The peak force coefficient $D(T)$ exhibits strong temperature dependence, modeled as:
\begin{equation}
\begin{aligned}
    D(T) = F_z \cdot \mu(T) = F_z \cdot \left( a_0 + a_1 T + a_2 T^2 + a_3 \exp(-a_4 T) \right),
    \end{aligned}
\end{equation}
where $\mu(T)$ is the temperature-dependent friction coefficient, and $a_0, \dots, a_4$ are empirical parameters determined through tire testing. This relationship captures the reduction in friction with increasing temperature due to rubber hysteresis effects.

\begin{figure}[h!]
    \centering
    \includegraphics[width=0.85\linewidth]{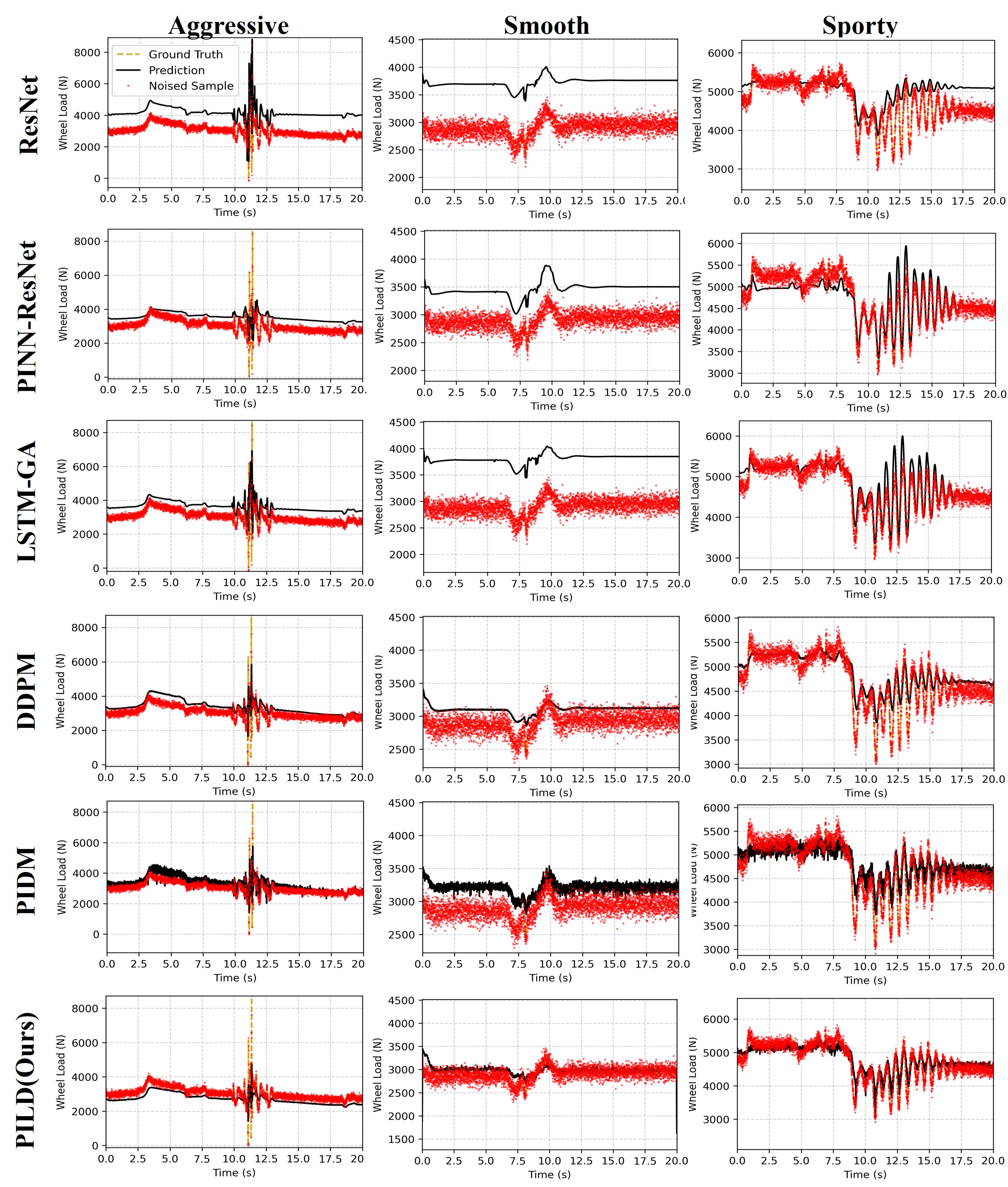}
    \caption{Experiment results on tire force estimation tasks.}
    \label{fig:result_wheelload}
\end{figure}

Similarly, the longitudinal peak coefficient $D_x(T)$ follows the temperature-dependent relationship:
\begin{equation}
    D_x(T) = F_z \cdot \left( c_0 + c_1 T + c_2 T^2 + c_3 \exp(-c_4 T) \right),
\end{equation}
where $c_0, \dots, c_4$ are empirical parameters for longitudinal friction, determined through tire testing.

To derive the partial differential equation (PDE) governing wheel load $F_z(t,T)$ (with $t$ as time and $T$ as measured tire temperature), we start by substituting the tire force expressions into the equilibrium equations and considering dynamic variations.

From the force equilibrium, resolve the three-dimensional force balance into scalar components:
\begin{equation}
    \sum_{i \in \{p,u1,u2,t,l1,l2\}} F_{i,x} = F_x(F_z, T, s), \
    \sum_{i \in \{p,u1,u2,t,l1,l2\}} F_{i,y} = F_y(F_z, T, \alpha), 
    \sum_{i \in \{p,u1,u2,t,l1,l2\}} F_{i,z} = F_z. 
\end{equation}

The left-hand sides of Equations of $F_x$ and $F_y$ depend on suspension geometry, which is a function of spring-damper compression $x_d(t)$. Using the chain rule, their time derivatives become:
\begin{subequations}
\begin{align}
    \frac{d}{dt}\left( \sum F_{i,x} \right) &= \sum \left( \frac{\partial F_{i,x}}{\partial x_d} \cdot \frac{\partial x_d}{\partial t} + \frac{\partial F_{i,x}}{\partial \dot{x}_d} \cdot \frac{\partial^2 x_d}{\partial t^2} \right), \label{eq:dxdt} \\
    \frac{d}{dt}\left( \sum F_{i,y} \right) &= \sum \left( \frac{\partial F_{i,y}}{\partial x_d} \cdot \frac{\partial x_d}{\partial t} + \frac{\partial F_{i,y}}{\partial \dot{x}_d} \cdot \frac{\partial^2 x_d}{\partial t^2} \right), \label{eq:dydt}
\end{align}
\end{subequations}
where $\dot{x}_d = \partial x_d/\partial t$ and $\ddot{x}_d = \partial^2 x_d/\partial t^2$ are the velocity and acceleration of the spring \& damper assembly, respectively.

The right-hand sides of Equations of $F_x$ and $F_y$ depend on $F_z$ and $T$ (measured as $T(t)$). Their total derivatives with respect to time are:
\begin{equation}
    \frac{dF_x}{dt} = \frac{\partial F_x}{\partial F_z} \cdot \frac{\partial F_z}{\partial t} + \frac{\partial F_x}{\partial T} \cdot \frac{dT}{dt}, \quad
    \frac{dF_y}{dt} = \frac{\partial F_y}{\partial F_z} \cdot \frac{\partial F_z}{\partial t} + \frac{\partial F_y}{\partial T} \cdot \frac{dT}{dt}, \label{eq:dfydt}
\end{equation}
where $\partial F_x/\partial F_z$ and $\partial F_x/\partial T$ are partial derivatives of the longitudinal force with respect to vertical load and temperature, derived from the Pacejka Magic Formula (Equations \eqref{e27}).

Equating the time derivatives from suspension dynamics and tire forces, then solving for $\partial F_z/\partial t$ yields the coupled PDE:
\begin{equation}
\begin{aligned}
    \frac{\partial F_z(t,T)}{\partial t} = \Bigg[ \left( \frac{\partial F_x}{\partial F_z} \right)^2 + \left( \frac{\partial F_y}{\partial F_z} \right)^2 \Bigg]^{-1} 
    \cdot \Bigg\{ \frac{\partial F_x}{\partial F_z} \left[ \sum \left( \frac{\partial F_{i,x}}{\partial x_d} \dot{x}_d + \frac{\partial F_{i,x}}{\partial \dot{x}_d} \ddot{x}_d \right) - \frac{\partial F_x}{\partial T} \dot{T} \right] 
    \\+ \frac{\partial F_y}{\partial F_z} \left[ \sum \left( \frac{\partial F_{i,y}}{\partial x_d} \dot{x}_d + \frac{\partial F_{i,y}}{\partial \dot{x}_d} \ddot{x}_d \right) - \frac{\partial F_y}{\partial T} \dot{T} \right] \Bigg\},
\end{aligned}
\end{equation}
where $\dot{T} = dT/dt$ is the time rate of change of measured tire temperature.

This PDE explicitly captures:
(1) \textbf{Time dependence} through spring-damper dynamics ($x_d, \dot{x}_d, \ddot{x}_d$) from suspension motion,
(2) \textbf{Temperature dependence} through measured $T$ and its rate $\dot{T}$, which modify tire friction coefficients,
(3) \textbf{Coupling} between vertical load $F_z$ and horizontal forces $F_x, F_y$ via the Pacejka model derivatives.

Therefore, there are only six unknown quantities left in the entire system (force magnitudes $F_p, F_{u1}, F_{u2}, F_t, F_{l1}, F_{l2}$). With the six-dimensional equilibrium equations for force balance and moment balance (Equation \eqref{eq24}), combined with the derived PDE for $F_z(t,T)$, the problem is ultimately solved.

\begin{table}[htbp!]
\begin{center}
\setlength{\tabcolsep}{15pt} 
\footnotesize

{
\begin{tabular}{l|c|c|c}
\toprule
\multirow{2}*{\textbf{Method}} & \multicolumn{1}{|c|}{\textbf{Aggressive}} & \multicolumn{1}{|c}{\textbf{Smooth}} & \multicolumn{1}{|c}{\textbf{Sporty}} \\ 
~ & \textbf{$e_{F}$$\downarrow$}          & \textbf{$e_F$$\downarrow$}          & \textbf{$e_F$$\downarrow$}   \\

\midrule

EKF &  1008.127$\pm$213.478&  597.363$\pm$89.737 & 716.543$\pm$95.104 \\

FCN &  1257.633$\pm$221.739&  601.559$\pm$98.712 & 794.780$\pm$102.563 \\

ResNet & 1121.367$\pm$204.319 & 585.561$\pm$79.301 &647.976$\pm$98.417 \\

PINN-FCN & 1012.849$\pm$205.850 & 551.508$\pm$81.638 & 654.644$\pm$110.379 \\

PINN-ResNet  &994.783$\pm$178.662 &642.110$\pm$78.361 & 653.944$\pm$95.558 \\

B-PINN  &1002.531$\pm$153.297 &598.617$\pm$80.336& 661.279$\pm$91.676 \\

LSTM-RNN & 1302.589$\pm$228.605 & 720.567$\pm$132.814 & 788.163$\pm$140.387 \\
LSTM-GA & 1150.434$\pm$219.348 & 582.840$\pm$96.622 & 642.638$\pm$102.228\\

DDPM & \underline{980.328$\pm$142.894} & \underline{560.796$\pm$63.707} & 660.897$\pm$75.630\\

PIDM & 988.642$\pm$150.277 & 579.546$\pm$57.095 & \underline{630.543$\pm$85.367} \\

\textbf{PILD} (Ours) &  \textbf{958.578$\pm$146.653} &  \textbf{520.631$\pm$51.683} & \textbf{607.274$\pm$92.531} \\ \bottomrule

\end{tabular}
\caption{Quantitative results with sample-level standard deviations on tire force estimation tasks. The $\pm$ values are computed across generated samples in a single trained-model evaluation.}
\label{table:app:tire}
}

\end{center}
\end{table}

To verify the generalization ability across different vehicle models, we shuffled data from various vehicle types to construct the training and test datasets.
More experimental results are demonstrated in Figure \ref{fig:result_wheelload}.
We also performed statistical analysis to determine data magnitude and applied fixed scaling factors during computation, then reversed the scaling afterward to normalize the data.
The metric here is the deviation in tire load (N).

\subsubsection{Darcy Flow}
\label{darcy}

Darcy flow describes the seepage behavior of fluids in porous media and serves as a canonical benchmark for validating physics-constrained generative models. This section presents an overview of its core theoretical components, including the governing equations, numerical solution approach, and dataset construction.

The flow of incompressible fluids through porous media is governed by Darcy's law and the principle of mass conservation, which together form the following partial differential equation (PDE):
\begin{equation}
-\nabla \cdot \left( K(\boldsymbol{x}) \nabla p(\boldsymbol{x}) \right) = f_s(\boldsymbol{x}), \quad \boldsymbol{x} \in \Omega \subset \mathbb{R}^2,
\end{equation}
where $p(\boldsymbol{x})$ is the pressure field, $K(\boldsymbol{x})$ is the spatially-varying permeability field that characterizes the medium's conductivity, and $f_s(\boldsymbol{x})$ is a prescribed source/sink term (modeling fluid injection or extraction). The domain $\Omega$ is typically taken as the unit square $[0,1]^2$. This system is subject to homogeneous Neumann boundary conditions ($\nabla p \cdot \boldsymbol{n} = 0$ on $\partial \Omega$), representing no-flow boundaries, and a zero-mean pressure constraint ($\int_{\Omega} p(\boldsymbol{x}) \, d\boldsymbol{x} = 0$) to ensure solution uniqueness.

The fluid velocity field $\boldsymbol{u}(\boldsymbol{x})$ is related to the pressure gradient via Darcy's law:
\begin{equation}
\boldsymbol{u}(\boldsymbol{x}) = -K(\boldsymbol{x}) \nabla p(\boldsymbol{x}).
\end{equation}
This constitutive relation implies that fluid flows from regions of high pressure to low pressure, with the local flux being modulated by the permeability of the medium at that point. In practice, the velocity field is often not directly used as a model output; instead, it can be post-processed from the pressure solution using finite difference approximations of the gradient.

\begin{figure}[htbp]
    \centering
    \includegraphics[width=0.65\linewidth]{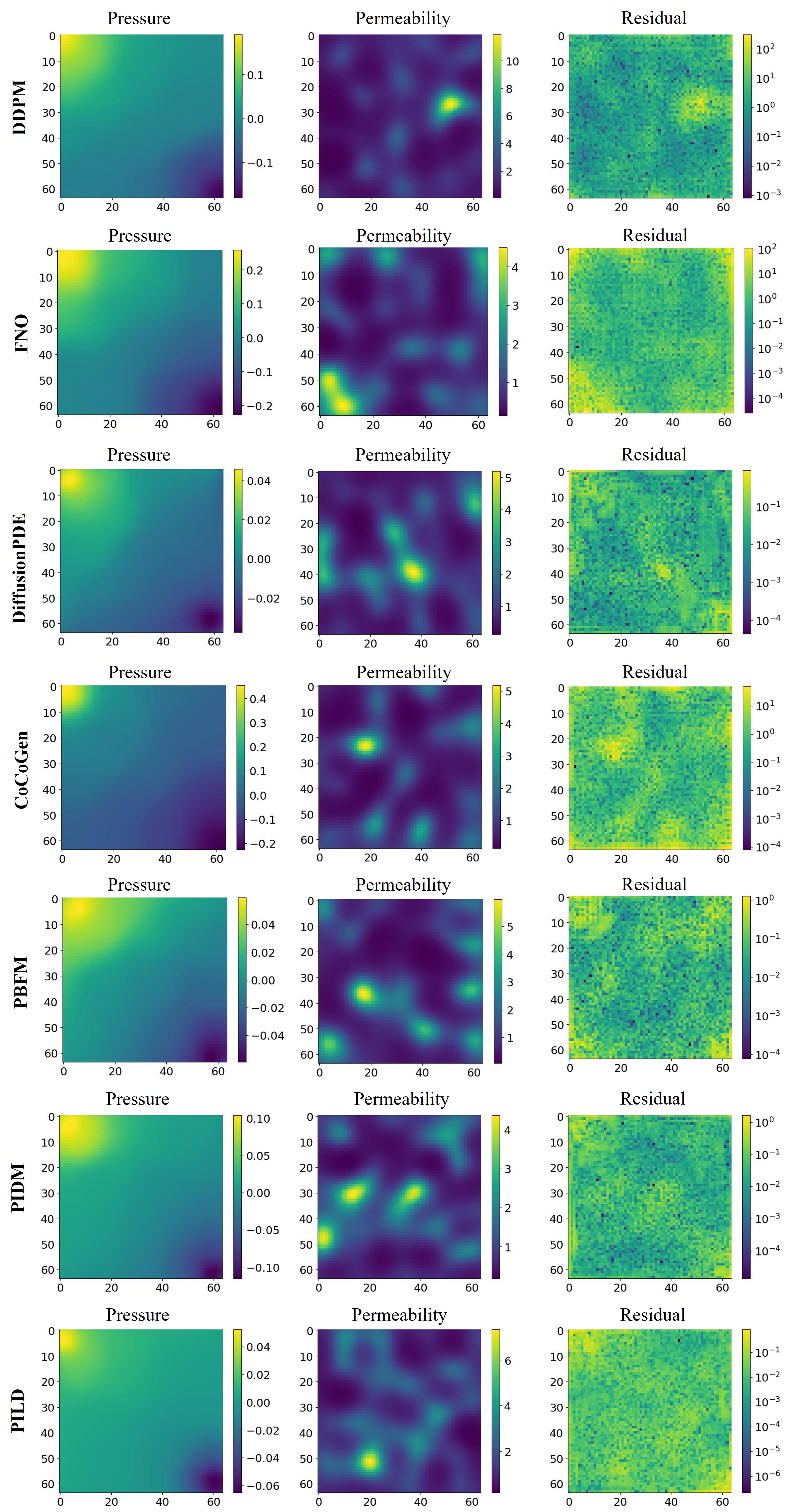}
    \caption{Experiment results on Darcy flow generation.}
    \label{fig:darcy2}
\end{figure}

In the standard benchmark setup adopted from prior work such as CoCoGen~\cite{jacobsen2025cocogen}, the source term $f_s(\boldsymbol{x})$ is fixed to a constant value across all samples (e.g., $f_s \equiv 1$), thereby isolating the effect of the heterogeneous permeability field $K(\boldsymbol{x})$ on the resulting pressure distribution. The permeability field itself is generated stochastically: it is obtained by applying an exponential transform to a random field with a specified correlation length (typically $\ell = 0.1$ or $0.2$). This ensures that $K(\boldsymbol{x})$ remains strictly positive (a physical requirement) while exhibiting spatially correlated structures that mimic real geological formations.

To construct the dataset, the continuous PDE is discretized on a uniform Cartesian grid of size $s \times s$, where $s=64$ is the resolution. At each grid node $(i,j)$, the spatial coordinate is given by $\boldsymbol{x}_{i,j} = \big((i-1)/(s-1), (j-1)/(s-1)\big)^\top$. A linear system $\boldsymbol{A}\boldsymbol{p} = \boldsymbol{f}$ is then assembled using second-order finite differences to approximate the divergence and gradient operators in the PDE. Solving this sparse linear system yields the discrete pressure values $\{p_{i,j}\}$ at all grid points, forming the field $p(\boldsymbol{x})$.

Consequently, each training instance consists of a paired sample $\{K(\boldsymbol{x}), p(\boldsymbol{x})\}$, where both fields are represented as $s \times s$ images. The mapping from permeability to pressure is highly non-linear and non-local, making it a challenging inverse problem and an excellent testbed for evaluating a generative model’s ability to capture complex physical dependencies. 

In this experiment, both the horizontal and vertical coordinates range from 0 to 64, corresponding to 64 pixels. The training and test datasets are generated via numerical simulation.
Figure \ref{fig:darcy2} demonstrates more samples of the experiment.
Following \cite{jacobsen2025cocogen}, no additional normalization is needed as both permeability and pressure fields are naturally around order 1.
The metric is RMAE.

\subsubsection{Plasma Dynamics}
\label{plasma}

Low-temperature, magnetized plasmas in the edge region of tokamaks (e.g., scrape-off layer, pedestal) exhibit inherently complex turbulent behavior driven by instabilities such as drift waves, resistive interchange modes, and blob propagation \cite{greenwald201420,labombard2005evidence}, which govern particle and energy transport, wall loading, and pedestal stability—critical factors for the viability of magnetic confinement fusion reactors. A widely adopted theoretical framework for simulating this regime is the drift-reduced Braginskii model \cite{fitzpatrick2022plasma}, which provides a set of coupled fluid equations describing the evolution of key plasma quantities (density, temperature, velocity, electrostatic potential) while retaining essential physics of magnetized collisional plasmas. This work establishes a benchmark model focused on the dynamics of electron density $n_e(\boldsymbol{x}, t)$ and electron temperature $T_e(\boldsymbol{x}, t)$—fields directly measurable via diagnostics (e.g., gas puff imaging \cite{zweben2017invited}, Langmuir probes) and central to characterizing edge turbulence—by deriving the normalized two-dimensional (2D) model from the full 3D drift-reduced Braginskii equations, detailing key approximations, and specifying numerical parameters consistent with tokamak edge experiments.

Here we have to note that using simulated data is the established paradigm for edge plasma turbulence research. Experimental plasma diagnostics provide only partial observations with limited spatial-temporal resolution. Consequently, virtually most prior works \cite{mathews2021uncovering,toscano2025pinns} on data-driven plasma modeling rely on synthetic data. Our choice is consistent with this community standard. However, if real data becomes available, we will extend our evaluation accordingly. Also, the benchmark's complexity makes it a meaningful testbed. The governing PDEs are highly nonlinear and coupled, making this a challenging system, and all baselines were evaluated under the same setting, ensuring that the comparison is fair and reflects methodological differences.

The drift-reduced Braginskii model is derived under core assumptions including the low-$\beta$ condition (plasma pressure negligible compared to magnetic pressure, justifying the electrostatic approximation $\mathbf{B} = B_0 \hat{\mathbf{z}}$), magnetized species (ion gyration frequency $\Omega_i \gg$ turbulent fluctuation frequency $\partial/\partial t$, allowing neglect of ion inertia in perpendicular dynamics), quasineutrality ($n_e \approx n_i = n$, no local charge imbalance with $\nabla \cdot \mathbf{j} = 0$), and the collisionless limit for edge plasmas (collisional drifts and viscosity negligible compared to drift-wave dynamics, except for diffusive losses). The 3D evolution equations for electron density $n_e$ and electron temperature $T_e$ (in physical units) are given by:
\begin{align}
\frac{d^e n_e}{dt} &= -\frac{2c}{B_0} \left[ n_e C_{(\phi)} - \frac{1}{e} C_{(p_e)} \right] - n_e \nabla_\parallel v_{\parallel e} + S_n + \mathcal{D}_{n_e}, \\
\frac{d^e T_e}{dt} &= -\frac{5}{3} \frac{T_e}{n_e} \nabla_\parallel v_{\parallel e} + \frac{2}{3n_e} \nabla_\parallel \left( \kappa_\parallel^e \nabla_\parallel T_e \right) + \frac{1}{n_e} S_{E,e} + \mathcal{D}_{T_e},
\end{align}
where $d^e f/dt = \partial_t f + (c/B_0)[\phi, f] + v_{\parallel e}\nabla_\parallel f$ is the species-convective derivative for electrons accounting for $\mathbf{E} \times \mathbf{B}$ drift ($\mathbf{E} = -\nabla \phi$) and parallel advection, $[\phi, f] = b_0 \times \nabla \phi \cdot \nabla f$ is the Poisson bracket (2D projection of $\mathbf{E} \times \mathbf{B}$ drift advection with $b_0 = \hat{\mathbf{z}}$), $C_{(f)} = b_0 \times \kappa \cdot \nabla f$ is the curvature operator ($\kappa = -\hat{R}/R$, tokamak major radius $R$, negligible in slab geometry for benchmarking), $p_e = n_e T_e$ is electron pressure (Boltzmann relation for ideal plasma), $\nabla_\parallel = \partial/\partial z$ is the parallel gradient, $v_{\parallel e}$ is parallel electron velocity, $\kappa_\parallel^e$ is parallel thermal conductivity, $S_n$ ($\text{m}^{-3}\text{s}^{-1}$) and $S_{E,e}$ ($\text{eV}\text{m}^{-3}\text{s}^{-1}$) are particle and energy sources, and $\mathcal{D}_{n_e}$ and $\mathcal{D}_{T_e}$ are diffusive terms (cross-field and parallel diffusion, dominated by anomalous transport in edge plasmas).

To simplify numerical implementation and ensure physical consistency, all variables are normalized using reference scales derived from tokamak edge parameters: reference density $n_0 = 5 \times 10^{19}\ \text{m}^{-3}$, reference temperature $T_0 = 25\ \text{eV}$, reference potential $\phi_0 = T_0/e$, reference length $a_0 = 0.22\ \text{m}$, and reference time $t_0 = a_0 / v_{E0}$ (with $v_{E0} = \phi_0/(B_0 a_0)$ as characteristic $\mathbf{E} \times \mathbf{B}$ drift velocity, $B_0 = 5\ \text{T}$). Normalized quantities are defined as:
\begin{align*}
\tilde{n}_e = \frac{n_e}{n_0}, \quad \tilde{T}_e = \frac{T_e}{T_0}, \quad \tilde{\phi} = \frac{\phi}{\phi_0}, \quad \tilde{\boldsymbol{x}} = \frac{\boldsymbol{x}}{a_0}, \quad \tilde{t} = \frac{t}{t_0},
\end{align*}
with normalized operators $\tilde{\nabla} = a_0 \nabla$, $\tilde{\nabla}_\perp^2 = a_0^2 \nabla_\perp^2$, and $[\tilde{f}, \tilde{g}] = a_0^2 [f, g]_{\text{physical}}$. Substituting these into the 3D equations and simplifying---neglecting parallel advection terms due to slow $z$-direction variation ($\tilde{\nabla}_\parallel \to 0$, 2D slab approximation), curvature terms in slab geometry, approximating diffusive terms as anomalous cross-field diffusion, and adding damping terms to model particle/energy losses to the wall---yields the normalized 2D equations for $\tilde{n}_e$ and $\tilde{T}_e$:
\begin{align}
\frac{\partial \tilde{n}_e}{\partial \tilde{t}} + [\tilde{\phi}, \tilde{n}_e] &= D_n \tilde{\nabla}_\perp^2 \tilde{n}_e - \alpha_n (\tilde{n}_e - 1) + \tilde{S}_n, \label{eq:ne_eq_normalized} \\
\frac{3}{2} \frac{\partial \tilde{T}_e}{\partial \tilde{t}} + \frac{3}{2} [\tilde{\phi}, \tilde{T}_e] + \frac{1}{2} [\tilde{\phi}, \tilde{n}_e] &= D_T \tilde{\nabla}_\perp^2 \tilde{T}_e - \alpha_T (\tilde{T}_e - 1) + \tilde{S}_T. \label{eq:Te_eq_normalized}
\end{align}

In these equations, the time derivative terms capture the temporal evolution of density and temperature, with the factor $3/2$ for temperature originating from the thermal energy of a monatomic gas ($U = \frac{3}{2}nT$). The $\mathbf{E} \times \mathbf{B}$ advection terms, defined via the Poisson bracket $[\tilde{\phi}, \tilde{f}] = \partial_x \tilde{\phi} \partial_y \tilde{f} - \partial_y \tilde{\phi} \partial_x \tilde{f}$, describe turbulent transport driven by the electric drift velocity, while the coupled advection term $\frac{1}{2}[\tilde{\phi}, \tilde{n}_e]$ in the temperature equation accounts for energy transport correlated with density gradients (derived from parallel heat flux closure). The anomalous diffusion terms $D_n \tilde{\nabla}_\perp^2 \tilde{n}_e$ and $D_T \tilde{\nabla}_\perp^2 \tilde{T}_e$ model cross-field transport from turbulent eddies, with $D_n$ and $D_T$ as anomalous diffusion coefficients, and the wall damping terms $-\alpha_n (\tilde{n}_e - 1)$ and $-\alpha_T (\tilde{T}_e - 1)$ enforce linear relaxation to equilibrium (representing particle and energy losses to tokamak walls). The source terms $\tilde{S}_n$ and $\tilde{S}_T$ correspond to external particle injection (e.g., gas puff) or energy input (e.g., radiofrequency heating).

\begin{figure}[htpb!]
    \centering
    \includegraphics[width=0.7\linewidth]{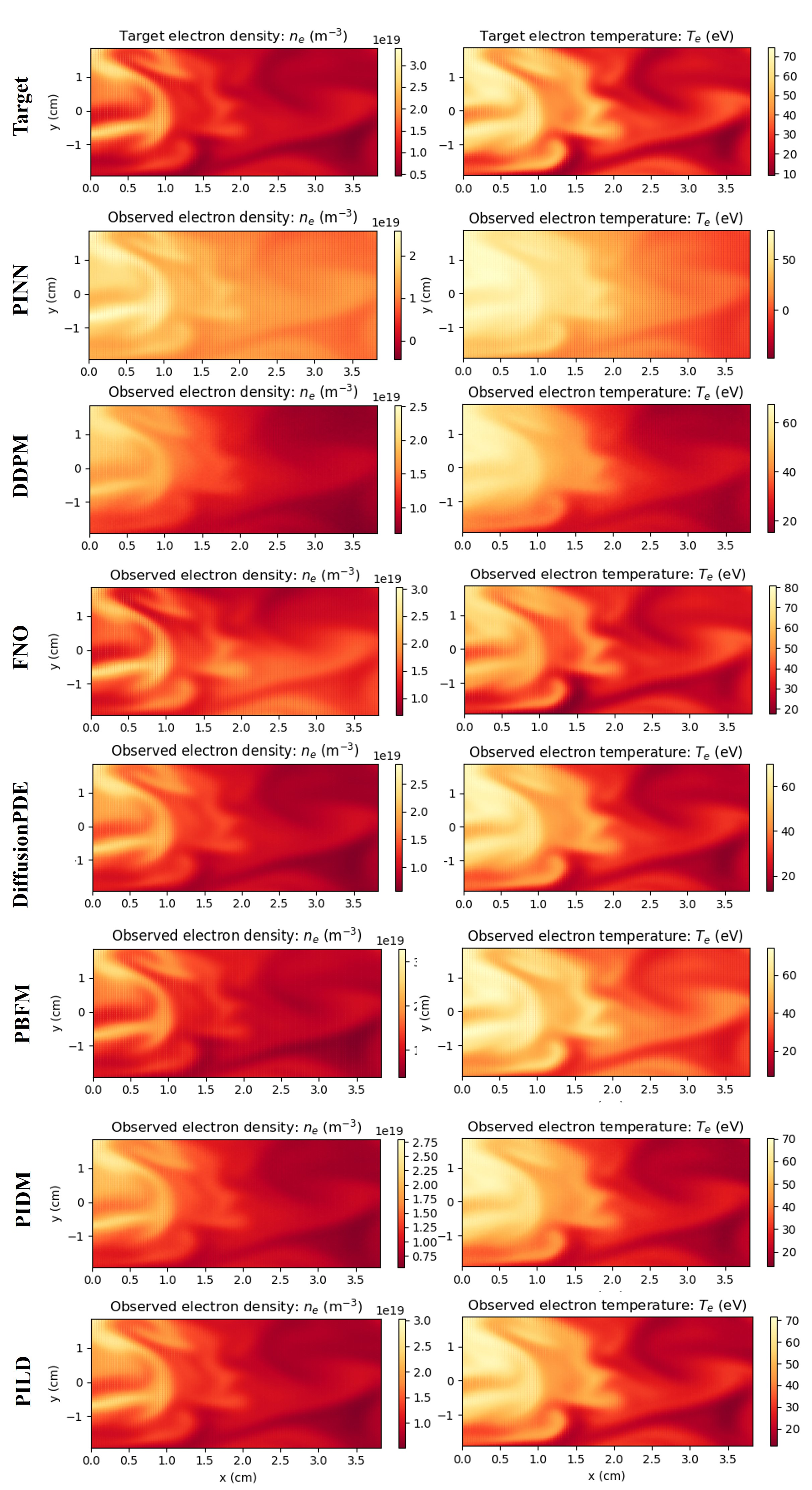}
    \caption{Experiment results on plasma dynamics prediction.}
    \label{fig:plasma}
\end{figure}

The electrostatic potential $\tilde{\phi}$ is determined self-consistently via quasi-neutrality, with the simplified Boltzmann approximation $\tilde{\phi} \approx \tilde{n}_e$ adopted for the benchmark model—valid for low-temperature edge plasmas and sufficient to close the system. For higher fidelity, the full quasi-neutrality condition $\nabla_\perp^2 \tilde{\phi} = \tilde{n}_e - \tilde{n}_i$ can be employed, though the Boltzmann approximation adequately captures core turbulent dynamics such as blob propagation.

The numerical setup for synthetic dataset generation is consistent with the global drift-ballooning (GDB) code and tokamak experiments: the 2D poloidal simulation domain spans physical dimensions $x \in [0, 3.8\ \text{cm}]$ and $y \in [0, 3.8\ \text{cm}]$ (normalized $\tilde{x}, \tilde{y} \in [0, 0.173]$) with grid resolution $\Delta \tilde{x} \approx 1.36 \times 10^{-3}$ and $\Delta \tilde{y} \approx 1.82 \times 10^{-3}$. Boundary conditions include homogeneous Neumann conditions for $\tilde{n}_e$ and $\tilde{T}_e$ (no net flux at domain edges) and a Dirichlet condition $\tilde{\phi} = 0$ at boundaries (enforcing radial $\mathbf{E} \times \mathbf{B}$ drift to zero). Initial conditions are truncated Gaussian profiles:
\begin{align*}
\tilde{n}_e(\tilde{x}, \tilde{y}, 0) = 1 + 0.2 \exp\left(-\frac{\tilde{x}^2 + \tilde{y}^2}{2\sigma^2}\right), \quad \tilde{T}_e(\tilde{x}, \tilde{y}, 0) = 1 + 0.15 \exp\left(-\frac{\tilde{x}^2 + \tilde{y}^2}{2\sigma^2}\right),
\end{align*}
where $\sigma = 0.05$ (normalized width, corresponding to $\approx 1\ \text{cm}$ physical width). Benchmark parameters calibrated to edge plasma turbulence are $D_n = 1.0 \times 10^{-4}$, $D_T = 1.5 \times 10^{-4}$, $\alpha_n = 0.1\ \tilde{t}^{-1}$, $\alpha_T = 0.08\ \tilde{t}^{-1}$, $\tilde{S}_n = 0.05$, and $\tilde{S}_T = 0.03$.

The system is solved using a second-order finite difference method with a trapezoidal leap-frog time integration scheme, implicit integration of diffusion terms (to avoid numerical instability from stiff diffusion), and a normalized time step $\Delta \tilde{t} = 1.0 \times 10^{-3}$ (corresponding to $\approx 4.55 \times 10^{-11}\ \text{s}$ physical time, consistent with the GDB code). This setup reproduces key edge plasma phenomena observed in experiments and full 3D simulations \cite{mathews2021uncovering}, including blob propagation (filamentary structures advected radially outward via $\mathbf{E} \times \mathbf{B}$ drift), turbulent transport (anomalous diffusion coefficients matching measured cross-field transport levels), and coupled density-temperature dynamics.

For PINNs, the residuals of Eqs. \eqref{eq:ne_eq_normalized} and \eqref{eq:Te_eq_normalized} serve as regularization to ensure learned models satisfy physical constraints even with partial or noisy observations:
\begin{align*}
\mathcal{R}_{\tilde{n}_e} &= \frac{\partial \tilde{n}_e}{\partial \tilde{t}} + [\tilde{\phi}, \tilde{n}_e] - D_n \tilde{\nabla}_\perp^2 \tilde{n}_e + \alpha_n (\tilde{n}_e - 1) - \tilde{S}_n, \\
\mathcal{R}_{\tilde{T}_e} &= \frac{3}{2} \frac{\partial \tilde{T}_e}{\partial \tilde{t}} + \frac{3}{2} [\tilde{\phi}, \tilde{T}_e] + \frac{1}{2} [\tilde{\phi}, \tilde{n}_e] - D_T \tilde{\nabla}_\perp^2 \tilde{T}_e + \alpha_T (\tilde{T}_e - 1) - \tilde{S}_T.
\end{align*}
The benchmark can be extended to toroidal geometry by reinstating curvature terms $C_{(f)} = -\frac{1}{R_0} \partial_y f$ (where $R_0 = 0.68\ \text{m}$ is the tokamak major radius \cite{greenwald201420}) and modifying the Poisson bracket to:
\begin{align*}
[\tilde{\phi}, \tilde{f}]_{\text{toroidal}} = \partial_x \tilde{\phi} \partial_y \tilde{f} - \partial_y \tilde{\phi} \partial_x \tilde{f} - \frac{1}{R_0} \partial_y \tilde{\phi} \partial_y \tilde{f},
\end{align*}
which captures resistive interchange instabilities dominant in bad-curvature regions of tokamaks.

Due to the scarcity of plasma data, the experimental data in this case are also obtained through numerical simulation \cite{mathews2021uncovering}.
Figure \ref{fig:plasma} demonstrates some results of the experiment, and we present the quantitative results again in Table \ref{table:app_plasma}.
Given the extremely large magnitudes, we adopted non-dimensionalization using reference temperature and density values from plasma literature to avoid unit inconsistency when computing residuals \cite{mathews2021uncovering}.
The predicted quantities are electron density ($m^{-3}$) and temperature (eV).

\begin{table}[t!]
\begin{center}
\setlength{\tabcolsep}{20pt} 
\footnotesize
{
\begin{tabular}{l|c|c}
\toprule
\multirow{1}*{\textbf{Method}} & Density error ($10^{19}$) & Temperature error \\  
\midrule

PINN & 0.531 $\pm$ 0.095 & 5.292 $\pm$ 0.538\\
DDPM  & 0.122 $\pm$ 0.020 & 1.766 $\pm$ 0.284\\
FNO & 0.153 $\pm$ 0.046 & 1.609 $\pm$ 0.258\\
DiffusionPDE & 0.112 $\pm$ 0.014 & \underline{0.728 $\pm$ 0.067}\\

PBFM  & {0.127 $\pm$ 0.025} & 1.293 $\pm$ 0.116 \\
PIDM  & \underline{0.107 $\pm$ 0.012} & 0.831 $\pm$ 0.061 \\
\textbf{PILD} (Ours)   & \textbf{0.074 $\pm$ 0.011} & \textbf{0.451 $\pm$ 0.055} \\

 \bottomrule
\end{tabular}
\caption{Quantitative results on plasma dynamics prediction with sample-level standard deviations. The $\pm$ values are computed across generated samples in a single trained-model evaluation.}
\label{table:app_plasma}
}

\end{center}
\end{table}


\end{document}